\documentclass[a4paper, onecolumn, review]{elsarticle}

\usepackage[english]{babel}
\usepackage{amsmath}
\usepackage{amssymb}
\usepackage{amsthm}
\usepackage{graphics}
\usepackage{amsfonts}
\usepackage{natbib}
\usepackage{geometry}
\usepackage{txfonts}
\usepackage{hyperref}

\usepackage{graphicx} 
\graphicspath{ {./LM-Eexpabs_images/} }
\biboptions{authoryear}

\usepackage{graphicx}
\graphicspath{ {./LM-Eexpabs_images/} }
\usepackage{subcaption} 

\usepackage{enumitem} 

\usepackage{colortbl}

\usepackage{multirow} 

\usepackage{float} 

\usepackage{booktabs}

\usepackage{nccmath}

\usepackage{nicematrix} 
\usepackage{tikz}
\usetikzlibrary{fit}

\usepackage{pdflscape} 

\usepackage{lipsum}

\usepackage{caption}
\usepackage{algorithm}
\usepackage{algpseudocode}

\algrenewcommand{\algorithmiccomment}[1]{\hfill \textcolor{blue}{//} #1}
\newenvironment{breakablealgorithm}
  {
     \refstepcounter{algorithm}
     \hrule height.8pt depth0pt \kern2pt
     \renewcommand{\caption}[2][\relax]{
       {\raggedright\textbf{Algorithm~\thealgorithm} ##2\par}%
       \ifx\relax##1\relax 
         \addcontentsline{loa}{algorithm}{\protect\numberline{\thealgorithm}##2}%
       \else 
         \addcontentsline{loa}{algorithm}{\protect\numberline{\thealgorithm}##1}%
       \fi
       \kern2pt\hrule\kern2pt
     }
  }{
     \kern2pt\hrule\relax
  }

\let\today\relax
\makeatletter
\def\ps@pprintTitle{%
    \let\@oddhead\@empty
    \let\@evenhead\@empty
    \def\@oddfoot{\footnotesize\itshape
         {Preprint} \hfill\today}%
    \let\@evenfoot\@oddfoot
    }
\makeatother

\begin{document}
	\begin{frontmatter}
		\title{Self-Adaptive, Dynamic, Integrated Statistical and Information Theory Learning}
		\author[1,2]{Zsolt János Viharos}
		\ead[1]{viharos.zsolt@sztaki.hu} 
		\author[1]{Ágnes Szűcs}
		\address[1]{Intitute for Computer Science and Control, Center of Excellence in Production Informatics and Control, Eötvös Loránd Research Network (ELKH), Centre of Excellence of the Hungarian Academy of Sciences (MTA), H-1111, Budapest, Kende u. 13-17., Hungary}
		\address[2]{John von Neumann University, H-6000, Kecskemét, Izsáki u. 10., Hungary}
		
		\begin{abstract} 
			The paper analyses and serves with a positioning of various "error" measures applied in neural network training and identifies that there is no "best of" measure, although there is a set of measures with changing superiorities in different learning situations. An outstanding, remarkable measure called $E_{Exp}$ published by Silva and his research partners represents a research direction to combine more measures successfully with fixed importance weighting during learning. The main idea of the paper is to go far beyond and to integrate this relative importance into the neural network training algorithm(s) realized through a novel error measure called $E_{ExpAbs}$. This approach is included into the Levenberg-Marquardt training algorithm, so, a novel version of it is also introduced, resulting a self-adaptive, dynamic learning algorithm. This dynamism does not has positive effects on the resulted model accuracy only, but also on the training process itself. The described comprehensive algorithm tests proved that the proposed, novel algorithm integrates dynamically the two big worlds of statistics and information theory that is the key novelty of the paper.
		\end{abstract}
	
	
	\end{frontmatter}
    
	
	\section{Introduction} \label{sec:intro}
	
	Nowadays, Artificial Intelligence (AI) and Machine Learning (ML) have important roles in the several fields of life, many experts deal with their improvements and applications (\cite{viharos_reinforcement_2021}, \cite{viharos_survey_2015}). At the end of the previous century researchers have already tried to improve them with various theoretical inventions (\cite{prieto2016neural}),
	e.g. at modeling assignments by Artificial Neural Networks (ANNs) novel error metrics (for instance from the field of information theory) were already applied in the training algorithms, beyond the classical Mean-Squared Error (MSE). The next paragraphs reviews and details these key development steps over time, including their applicability and modeling behaviours.
	
	Among the first researches in this field, the paper of \cite{hopfield1987learning} introduced entropy as cost function in two types of neural networks: three-layer analog perceptron and three-layer Boltzmann network. Entropy was used because of its frequent application in the comparison of probability functions. This research compared the neural networks, but did not deal with the performance investigation of different error measures. The paper of \cite{watrous1992comparison} has already compared two error functions (squared error and relative entropy) in various optimization algorithms, although it did not conclude a clear similarity or difference between them.

	\cite{park1995information} investigated other alternatives for squared error (similar to the MSE) and relative entropy (similar to Cross-Entropy, CE) in a gradient descent training algorithm. The paper addressed the Kullback - Leibler (KL), Jensen (J) and Jensen-Shannon (JS) divergences as error measures as well. The functioning of the neural network was detailed and the behaviour of measures were analyzed with different network parameter settings in the backpropagation algorithm. The newly applied error metrics had some advantages against squared error and relative entropy, however, it was strongly dependent on the network parameters.

	The ANN training can be considerd  as an optimization method, in the work of \cite{erdogmus2002error} theoretical conclusions were proved for optimizing algorithms with information theory measures, which is an especially important aspect for the current research presented in this manuscript.
	In their paper the possibility of using the Shannon Entropy (SE) criterion in artificial neural network training was investigated. They proved that the minimization of $\alpha$-order R\'enyi Entropy (RE) minimize the Csisz\'ar distance, so, for a special case, the minimization of SE (RE $\alpha \to 1$) minimize the Kullback-Leibler divergence of input-desired and input-output pairs' joint densities. They admitted entropy with the approximated density function (Parzen windowing method and Gaussian kernel) and gave the same global minimum as the actual error density. Based on this results, the application of SE as the basic information theory measure was possible in supervised neural network training.

	This basic information theory measure (SE) was also the candidate of \cite{silva2005neural} to change the commonly used cost function MSE.
	Gradient descent algorithm was applied in the algorithm with variable learning rate. It was observed on experimental way that the stability of the algorithm is highly sensitive on the smoothing parameter of the nonparametric kernel estimator. By comparing SE to MSE and CE it was concluded that SE can reach better performance according to test error and its standard deviation.
	
	\cite{rady2011shannon} compared MSE and SE in a detailed way. MSE is a good choice for training in case of linear system and Gaussian noise, however for non-linear systems and non-Gaussian noise it fails. SE is an alternative for MSE, which can eliminate this problem. MSE and SE were tested in the paper under different conditions (with various activation functions and learning rates), it was found that both MSE and SE have advantages. MSE increases the convergence speed, but the success of convergence is higher with SE (the manuscript notes other comparative results for the network parameters as well).
	
	Another frequently used Information Theory Measure (ITM) in ANN learning is Cross-Entropy (CE). The aim of \cite{kline2005revisiting}'s research was to compare the efficiency of CE and Squared Error (like MSE) if they are used for estimating posterior probabilities but not the minimization of the classification error. The comparison was taken with simulated data and the robustness of CE was found.
	
	Rényi Entropy (RE) is a generalized ITM with parameter alpha ($\alpha$). In the learning algorithms the most frequently applied RE variation is its second order one (RE2, $\alpha = 2$). In the paper of \cite{rady2011reyni} MSE and RE were compared on experimental way. The training algorithms were analysed in various aspects, it was concluded that MSE increases the speed of the convergence more than RE, however the convergence rate is higher in case of RE.
	
	In the article of \cite{Yu2019} autoencoders were evaluated by information theory measures. The usage of a matrix-based RE analogue function was an especially interesting idea among the very useful detailed theoretical considerations. With this approach, it was not anymore necessary to approximate explicit probability density function of the data. The theoretical parts of this paper were confirmed by experimental results, too.
	
	Mutual Information (MI) is an ITM similar to KL divergence. It is very useful in the practical AI problems, where comparison of probability density functions are required.
	E.g. in early diagnosis of Alzheimer's disease, it can be useful for the detection of coupling strength abnormality of multivariate neural series. This is possible with various methods that was presented in the article of \cite{Wen2019}. One of them is the multivariate permutation conditional mutual information. A novel and  useful method was developed, which is robust, effective and less computational intensive then another similar methods.
	
	In the paper of \cite{Kitazono2020} a method was presented for specifying regions, so called "complexes" in the brain network. The algorithm founds this subsystems based on its property, which is in relation to the information loss. MI is an adequate solution for this problem, because of its information theoretical considerations, so it resulted in a fast and effective Hierarchical Partitioning algorithm.
	
	Not only the well-known ITM measures exists in the literature, but there are also newly developed measures for various kinds of data science problems. \cite{silva2008data} constructed a measure called Exponential Error ($E_{Exp}$) for artificial neural network training which integrates two mathematical fields such as statistics and information theory through promising combination of various measures. This new measure is especially important for the current research, so it is presented more in details below, in Section \ref{Eexp}.
	
	\cite{amaral2013using} analyzed the behaviour of different cost functions for training auto-encoders. Cost functions like CE, Sum of Squared Error (SSE, like MSE) and exponential risk (EXP, $E_{Exp}$ for $\tau >0$) of \cite{silva2008data} were applied in their pre-training and fine-tuning phases. The analyzed classification experiments show that the SSE is the best for pre-training, CE and EXP hold on earlier than required because of the naturally applied early stopping method. The various cost function pairs did not have high influence on classification results, though it could be observed that EXP function in fine-tuning treats well the unbalanced data. The similar behaviour for EXP and CE was concluded based on the experimental results, too.
	
	\cite{silva2014classification} assessed the performance of MSE and some information theory measures (CE, SE, quadratic RE (RE2), error density at zero error (ZEDM), EXP) in details in their paper. Multilayer Perceptron was trained using these error functions in the classical backpropagation algorithm. The experiments were collected on 35 real-world datasets and were analyzed statistically. It was found that the ubiquitous MSE has better alternatives such as CE and EXP, and it was noted that the SE and RE2 are not useful in most of the tested datasets.
	
	In addition to the different error measures, the other components and parameters of ANN learning algorithm are significant, too.
	\cite{rimer2006cb3} introduced a novel model measure specialized for classification problems in order to perform better than MSE and CE in the given type of modeling tasks. A key component was that the introduced method modifies the model error measure during training as it sets dynamically output target values for each training pattern. They also formulated that using common differentiable metrics like MSE relies on the assumption that sample outputs are offset by inherent Gaussian noise, being normally distributed about a cluster mean and identified that other error metrics (e.g. CE) are more suited to classification problems. However, in their experiments (that were also validated on well-known benchmark datasets) it was observed that MSE and CE optimized network models resulted nearly identical results, so, only the accuracy of training with CE were presented for brevity in the comparison to the novel solution. As result, the introduced method had higher test accuracy and tighter mean confidence interval then CE without weight decay in the backpropagation (BP) algorithm on 10 of the 11 test datasets.

	In the current research it is an important aspect, which measures have been already used in the Levenberg-Marquardt (LM) training/optimization as cost function. A novel method was created by \cite{thevenaz2000optimization} for image recognition by using the composition of the KL divergence and the LM algorithm. The required elements for the integration were described in the paper such as derivatives of KL, vanishing in the Hesse matrix. Based on experimental results, it was concluded that the new optimizer method is more accurate and faster than the other image registration methods known in 2000.
	
	The alignment or registration of a pair of images is an operation required in many applications in various practical fields. The scientific work of \cite{dowson2007mutual} is close to the current paper because they introduced MI as one of the very popular information theory measures (that is in special, close relation to SE) in the LM training of artificial neural networks. MI measures the information shared between two signals, in the given application field it was calculated using the joint Probability Distribution Function (PDF) of the intensities (amplitudes) of the two images (signals). The authors appointed their motivation in relation to the given application field as MI is only slightly more expensive for computation than MSE but has several advantages, namely, it tolerates nonlinear relationships between the intensities in images and it is robust to noise, too. They exploited the behaviour of MI in the LM algorithm in the calculation of the Hessian matrix to make the learning efficient. It was proved in their vision oriented application that the introduced inverse-compositional MI based measure significantly outperformed the MSE based techniques considering model accuracy (with ~15 \%), speed and stability of the training process as well.
	
	Based on the work of \cite{thevenaz2000optimization},  \cite{panin2008mutual} developed among others a template tracking method. The method applied the MI measure being integrated in the LM optimizer. Comparison to similarity measure Sum of Squared Difference (like MSE) was presented in the paper, so, the new algorithm is robust and applicable to 3D textured objects.
	
	As it was shown above, there exists a lot of ITMs and a lot of possibilities to combine and modify the training algorithms' details and parameters. So, this two parts could be mixed together such as seen in the researches of authors Heravi and Hodtani \citep{heravi2018comparison, heravi2018does, heravi2018new, heravi2019new}.
	Significant and important groundwork results for the whole community were published by \cite{heravi2018does} with both theoretical and also simulation basis, where the theoretical results are more dominant. They clarified the relations among MSE, Minimum Error Entropy (MEE) (related to RE2) and Maximum Correntropy (CorrE), by appointing two key factors influencing the superiorities: the Signal-to-Noise Ratio (SNR) and the distance of the incorporated noise in the analysed data from the Gaussian distribution. The Kullback-Leibler (KL) divergence can be applied to measure this later factor. In case of a "simple" Gaussian noise distribution at low SNR values, the MSE outperforms MEE (and CorrE) but at high SNR values their performances are the same. But in case of Cauchy noise the information theory measures (MEE and CorrE) significantly outperform the MSE measure, where the difference is higher at low SNR values. Similar is the performance with Laplace noise distribution where the information theory measures (MEE and CorrE) are again significantly better than MSE in case if the so called scaled parameter of the Laplace distribution is high and they perform similarly when this parameter is really low. These scientific results represented also clearly that depending on the analysis circumstances, data regions, their collection methods, representativness, the incorporated noise, etc., the superiority of the various measures is fluctuating.

	By using of information theory measures it is possible to develop new algorithms on other fields of AI methods, too. The work of \cite{Yang2020} is a pioneering scientific research, which deals with correntropy based extreme learning machine for semi-supervised learning. Their aim was developing an algorithm, which is robust for non-Gaussian noise. In this research CorrE criterion based semi supervised extreme machine learning was used for solving the robustness requirement. The paper of \cite{He2020} deals with the field of Non-Negative Matrix Factorization. The researchers used maximum entropy in a novel way which resulted in the novel Maximum Entropy and Correlated  Non-Negative Matrix Factorization. Their experimental results showed that the framework presented in this research was able to exceed other state-of-the-art methods.
	
	Information theory measures are also important in the (close) future environment of the promising quantum information processing systems that became more and more important in theory and in practice. In such systems, currently, the level of noise is comparable to range of the system variables, so, it is an extremely difficult field to control. \cite{krisnanda_creating_2021} provided a versatile unified state preparation scheme based on a driven quantum network composed of randomly-coupled fermionic nodes. The output of such a system is then superposed with the help of linear mixing where weights and phases are trained in order to obtain the desired output quantum states. This mixing is then performed in order to maximize the relative entropy of discord between the output fermionic nodes.
	
	\begin{table}
		\begin{tabular}{@{}ll@{}}
			\toprule
			\multicolumn{2}{l}{Abbreviations} \\
			\midrule
			AI & Artificial Intelligence \\
			ML & Machin Learning \\
			ANN & Artificial Neural Networks\\
			BP & Back-Propagation \\
			ITM & Information Theory Measure \\
			MSE & Mean Squared Error \\
			CE & Cross-Entropy \\
			KL & Kullback-Leibler divergence \\
			J & Jensen divergence \\
			JS & Jensen-Shannon divergence\\
			SE & Shannon Entropy \\
			RE & R\'enyi Entropy \\
			RE2 & quadratic (second order) RE \\
			MI & Mutual Information \\
			$E_{Exp}$ & Exponential Error \\
			CorrE & Correntropy \\
			MEE & Minimum Error Entropy \\
			EXP & $E_{Exp}$ for $\tau > 0$ \\
			ZED(M) & Zero Error Density (Minimisation) \\
			RR & Recognition Rate \\
			\bottomrule
		\end{tabular}
		\caption{This Abbreviations are used in the manuscript.}
		\label{tab:abbr}
	\end{table}	
	
	This paragraph reviewed the variety of error measure and artificial neural network training combinations, the next one presents a sophisticated superiority analysis among these measures from multiple algorithmic viewpoints. The third paragraph introduces the directly preceding $E_{Exp}$ measure followed by a state-of-the-art positioning of the error measures and training algorithms. The actual scientific gap is appointed in the fifth paragraph and the novel introduced measure $E_{ExpAbs}$ in the next one for realizing a self-adaptive, integrated statistical and information theory learning. Its insertion into the Levenberg-Marquardt algorithm is precisely described and the related, comprehensive test and evaluation circumstances are detailed in the next section. The ninths paragraph proves the superiority of the introduced novel measure and the new training method considering calculation time, model accuracy, learning stability/robustness and recognition rate (in classification assignments). Conclusions and outlook frames the proposed scientific novelties and the manuscript is finalised by the acknowledgement and references paragraphs. Abbreviations frequently used in this article are listed in the Table \ref{tab:abbr}.
	
	\section{Superiority of training error measures} \label{sec:superiority}
	In the review of the related historical scientific and state-of-the-art research results given in the previous paragraph, MSE and the other information theory measures were analyzed based on their effects on the training algorithms' and the resulted models' behaviours. The three key and most frequently applied aspects of the evaluations are Accuracy, Speed and Stability/Robustness, so, these viewpoints are applied for positioning the various training error measures. The section describes the superiority (or similar) positions of the above referred measures in a structured way.
	
		\subsection{Superiority based on model Accuracy}
		The modelling accuracy of the final model is probably the most important viewpoint for determining the superiority of the applied measures.
		
			\subsubsection{Superiority of MSE}
			
			Compared to CE and $E_{Exp}$, the SSE (which is similar to MSE) based pre-training achieved the best layer-wise reconstruction performance in the training of an auto-encoder \cite{amaral2013using}.
			
			In case of J and JS error-metric the learning could reach similar result in sine function prediction to squared error (like MSE) metric (\cite{park1995information}).
			
			\cite{rimer2006cb3}' experiments were based on well-known benchmark datasets, it was observed that Sum-Squared Error (similar to MSE) and CE yielded nearly identical accuracy results.
			
			In case of "simple" Gaussian noise distribution at low Signal-to-Noise Ratio (SNR) values the MSE outperforms MEE (and CorrE) but at high SNR values their performances are the same \cite{heravi2018does}.
			
			The analysis of \cite{heravi2019new} shows that at high SNR the information theoretic and MSE based model error criteria act the same as CorrE.
			
			\subsubsection{Superiority of SE}
			
			Time Delay Neural Networks (TDNNs) trained with SE predict Mackey-Glass chaotic time series more accurate then MSE trained TDNNs \cite{erdogmus2002error}.
		
			The research of \cite{silva2005neural} found that SE has a better performance compared to MSE and CE according to misclassification error and its standard deviation.
			
			Good performance of Shannon Entropy and underperformance of MSE and ZED is found by \cite{silva2014classification}. It was observed based on measures, which were computed on balanced error rate quantities.
			
			\subsubsection{Superiority of CE}
			
			The results for the relative entropy (such as CE) show that relative entropy is able to predict sine function with such smaller error as squared error (such as MSE) \cite{park1995information}.
			
			In case of J and JS error-metric, it can reach similar result in sine function prediction than squared error (MSE) or CE \cite{park1995information}.
			
			\cite{rimer2006cb3} also described that using common differentiable metrics like MSE relies on the assumption that sample outputs are offset by inherent Gaussian noise, being normally distributed around a cluster mean and identified that other error metrics (e.g. CE) are more suited to classification problems.
			
			For auto-encoder training, CE has the lowest error on the first hidden layer compared to MSE and $E_{Exp}$ \cite{amaral2013using}.
			
			Based on statistical analysis by \cite{silva2014classification}, the underperformance of MSE is concluded compared to $E_{Exp}$ integrating CE as well.
			
			\subsubsection{Superiority of CorrE}
			
			In case of Cauchy noise, the information theory measures (MEE and CorrE) significantly outperformed the MSE measure, where this difference is higher at low SNR values. Similar is the performance in case of Laplace noise distribution, the information theory measures (MEE and CorrE) are again significantly better than MSE. It is valid when the so called scaled parameter of the Laplace distribution is high, and they performs similar when this parameter is very low \cite{heravi2018does}.
			
			In case of exponential noise when the SNR value is high, the information theory measures (MEE and CorrE) and the MSE served with the same model accuracy. At exponential noise, when the SNR ratio is low, the MEE and CorrE based training algorithms performs on the same level considering model accuracy, but both of them outperformed significantly the MSE based solution. Having Cauchy noise, the CorrE based solution has far the best accuracy and the MSE based neural network training algorithm performed as worst. In the concrete practical case of channel estimation, for the signal in the equalizer output, the results are the same as at the analysis with Cauchy noise, so, CorrE is the best one in accuracy aspect and MSE has the far lowest performance \cite{heravi2019new}.

			\subsubsection{Superiority of other measures}
			Good performance of 2-order R\'enyi entropy and underperformance of MSE and ZED is found by \cite{silva2014classification}. It was observed based on measures, which were computed on balanced error rate quantities.
			
			Based on statistical analysis by \cite{silva2014classification} the underperformance of MSE are concluded compared to the novel, introduced $E_{Exp}$.
			
			The scientific work of \cite{dowson2007mutual} introduced Mutual Information (MI) as a popular information theory measure (that is in special, close relation to Shannon Entropy (SE)) in training of artificial neural networks. It was proven in their vision oriented application that the introduced inverse-compositional MI based measure significantly outperformed the MSE based techniques considering model accuracy (by $
			\sim15\%$).

			J and JS error-metric can reach similar results in sine function prediction as the SE (such as MSE) (\cite{park1995information}).
			
			The Table \ref{table_accuracy} shows the number of superiorities of the individual error measures among each-other considering the resulted model accuracy.
			
			\begin{table}[ht!]
				\centering

				\begin{tabular}{|c||c|c|c|c|c|c|c|c|c|}
					\hline
					Accuracy & MSE & SE & CE & CorrE & RE(2) & MI & J & JS & $E_{Exp}$ \\
					\hline \hline
					MSE & \cellcolor{gray} & ++ & $\circ$ +++ $\circ$ & ++ $\circ$ & + & + & $\circ$ & $\circ$ & + \\\hline
					CE & $\circ$ + $\circ$ & + & \cellcolor{gray} & & & & $\circ$ & $\circ$ & \\\hline
					CorrE & + $\circ$ & & & \cellcolor{gray} & & & & & \\\hline
					RE(2) & & & & + & \cellcolor{gray} & & & & \\\hline
					ZED(M) & & + & & & + & & & & \\\hline
					J & $\circ$ & & $\circ$ & & & & \cellcolor{gray} & &  \\\hline
					JS & $\circ$ & & $\circ$ & & & & & \cellcolor{gray} & \\\hline
					$E_{Exp}$ & + & & + & & & & & & \\\hline \hline
					Number of measure applications & 8 & 4 & 8 & 4 & 2 & 1 & 2 & 2 & 1 \\\hline
				\end{tabular}
				\caption{Superiority of measures according to the final model accuracy. The headers of the columns are the various error measures according to the state-of-the-art scientific literature, while the rows represent the inferior measures. A '+' represents an important result in a publication that the measure according to the given column is superior to the measure according to the given row. Using of '$\circ$' shows that the measure in the given column and the measure in the given row reach similar result in the reported research. Sequences of '+'-s represent that several papers concluded in the same superior position. One can identify that MSE and CE are the most often applied measures.}
				\label{table_accuracy}
			\end{table}
		
		\subsection{Superiority based on training Speed}
		The applied modelling measure has significant effect on the training speed, too, consequently, the various measures need to be compared according to this criteria as well.
			
			\subsubsection{Superiority of MSE}
			
			The speed of the convergence is higher in case of MSE than with Shannon entropy for multilayer neural networks as concluded by \cite{rady2011shannon} after a detailed comparison of these two measures.
			
			The research of \cite{rady2011reyni} compared MSE and RE in the iteration number aspect. It was found that using RE needs more iteration than using MSE for all activation functions and for all learning rates of neural network models' trainings.
			
			In auto-encoder pre-training by \cite{amaral2013using}, the SSE (such as MSE) is slightly faster than CE. However the authors  additionally noted that the computation incorporates the cost itself not only its derivatives. As in the majority of scientific analyses, number of required steps using early stopping was applied to measure the speed of the convergence.
			
			In case of exponential noise when the SNR value is high, the MSE based algorithm converges much quicker compared to the information theory measures (MEE (similar to RE2) and CorrE) by \cite{heravi2019new}.
			
			\subsubsection{Superiority of CE}
			
			The neural network with relative entropy (CE) error-metric converges faster than the network with Square Error (MSE) for sine function predicting as reported by \cite{park1995information} 
						
			\subsubsection{Superiority of Correntropy}
			
			At exponential noise, when the SNR ratio is low, the MEE and CorrE based training algorithms perform on the same level, but both of them outperform significantly the MSE based solution considering the training speed. In the concrete practical case of channel estimation, CE was the best one in speed, MEE was the second and MSE had the far lowest performance reported by \cite{heravi2019new}.
						
			\subsubsection{Superiority of $E_{Exp}$}
			
			The work of \cite{silva2008data} observed good performance of the generalized $E_{Exp}$ considering the number of required training steps that is around three times less than with the other training measures (MSE, CE).
			
			In the auto-encoder pre-training $E_{Exp}$ is faster than CE \cite{amaral2013using}, similar to MSE related superiority.
			
			\subsubsection{Superiority of other measures}

			When \cite{dowson2007mutual} introduced MI as an information theory measure in training of artificial neural networks, they also identified that this measure significantly outperformed the MSE based techniques considering the speed of the training process.
			
			Using the J and JS metrics, rapid convergence was measured in case of sine prediction which was faster than (M)SE and CE \cite{park1995information}.
			
			At exponential noise, when the SNR ratio is low, the MEE (similar to RE2) and the CE based training algorithms perform similarly considering training speed, but both of them outperform significantly the MSE based solution. In the concrete practical case of channel estimation, CE was the best one according to speed, MEE was the second and MSE had the far lowest performance introduced in \cite{heravi2019new}.
			
			The Table \ref{table_speed} shows the number of superiorities of the individual model measures among each-other.
			
			\begin{table}[h]
				\centering
				\begin{tabular}{|c||c|c|c|c|c|c|c|c|}
					\hline
					Training speed & MSE & CE & CorrE & RE(2) & MI & J & JS & $E_{Exp}$ \\
					\hline \hline
					MSE & \cellcolor{gray} & + & + & + & + & + & + & + $\circ$ \\\hline
					SE & + & & & & & & &  \\\hline
					CE & ++ & \cellcolor{gray} & & & & + & + & ++ \\\hline
					CorrE & + & & \cellcolor{gray} & & & & & \\\hline
					RE(2) & ++ & & + & \cellcolor{gray} & & & & \\\hline
					$E_{Exp}$ & $\circ$ & & & & & & & \\\hline \hline
					Number of measure applications & 7 & 1 & 2 & 1 & 1 & 2 & 2 & 4 \\\hline
				\end{tabular}
				\caption{
				Superiority of measures above others according to the convergence speed of the learning algorithm. The headers of the columns are superior measures according to the state-of-the-art scientific literature, while the rows represent the inferior measures. A '+' represents an important result in a publication when the measure according to the given column is superior to the measure according to the given row. Using of '$\circ$' show that the measure in the given column and the measure in the given row serves with similar results. Sequences of '+'-s represent that several papers concluded in the same superior position. One can identify that MSE, CE and $E_{Exp}$ incorporating them are the most often occurring superior measures according training speed.}
				\label{table_speed}
			\end{table}
		
		\subsection{Superiority based on Stability/Robustness}
		Techniques introduced to improve the training speed naturally involve higher risk for divergence, consequently, the effects on the stability during learning using the different measures have to be analyzed and compared as well. It is in relation to the robustness of the learning algorithm against many other factors like initial weight values, unbalanced training data, outliers, noise, etc. Consequently, the stability and robustness are considered together, it is interesting that only a smaller ratio of scientific papers represent such analysis. However, a dedicated future research topic is possible to analyze the effect on separated stability and robustness in a much deeper level.
		
			\subsubsection{Superiority of SE}
			
			Entropy trained TDNN for fitting the distribution of a signal is able to disregard outliers (robustness) in Mackey-Glass chaotic time series and in non-linear system identification (better than MSE) as shown by \cite{erdogmus2002error}.
			
			In the SE-MSE comparison of \cite{rady2011shannon} it was observed that the extent of convergence is higher in case of SE.
			
			\subsubsection{Superiority of CE}
			
			Sine function prediction with KL metric has unstable convergence when compared to CE, MSE, J, JS by \cite{park1995information}.
			
			\cite{kline2005revisiting} found that CE in feed-forward neural networks is able to estimate posterior probabilities, and can outperform MSE in this aspect.
			
			\subsubsection{Superiority of Correntropy}
			
			At exponential noise, when the SNR ratio is low, the MEE and CE based training algorithms performed on the same level considering model training stability, but both of them outperform significantly the MSE based solution. In the concrete practical case of channel estimation, at the signal in the equalizer output, the results are the same as at the analysis with Cauchy noise, so, CE is the best one in stability, MEE is the second and MSE has the far lowest performance introduced by \cite{heravi2018new}.
			
			\subsubsection{Superiority of $E_{Exp}$}
			
			For small datasets it was observed that $E_{Exp}$ result similar or lower standard deviation than CE or MSE, so it depends less on train/test partition according to \cite{silva2008data}.
			
			Fine-tuning cost function pairs does not failed on a heavily unbalanced dataset applied by  auto-encoder pre-trainings, which was the fine-tuning function of $E_{Exp}$ in \cite{amaral2013using}.
			
			\subsubsection{Superiority of other measures}
			
			Sine prediction with KL metric had more unstable convergence compared to CE, MSE, J, JS measures by  \cite{park1995information}.
			
			In the R\'enyi entropy vs. MSE comparison of \cite{rady2011reyni} it was observed that the degree of convergence is higher in case of R\'enyi entropy.
			
			The results of \cite{dowson2007mutual} proved that the introduced inverse-compositional MI based measure significantly outperformed the MSE based techniques considering stability of the training process.
			
			The Table \ref{table_stability_robustness} shows the number of superiorities of the individual model measures among each-other considering stability/robustness.
			\begin{table}[h]
				\centering
				\begin{tabular}{|c||c|c|c|c|c|c|c|c|c|}
					\hline
					Stability & MSE & SE & CE & CorrE & RE(2) & MI & J & JS & $E_{Exp}$ \\
					\hline \hline
					MSE & \cellcolor{gray} & + + & + & + & + & + & & & ++ \\\hline
					SE & & \cellcolor{gray} & & + & & & & & \\\hline
					CE & & & \cellcolor{gray} & & & & & & + \\\hline
					KL & + & & + & & & & + & + & \\\hline \hline
					Number of measure applications & 1 & 2 & 1 & 2 & 1 & 1 & 1 & 1 & 3 \\\hline
				\end{tabular}
				\caption{Superiority of measures above others according to the stability/robustness of the learning process. The headers of the columns are superior measures according to the state-of-the-art scientific literature, while the rows represent the inferior measures. A '+' represents an important result in a publication that the measure according to the given column is superior to the measure according to the given row. Using of '$\circ$' show that the measure in the given column and the measure in the given row serves with similar results. Sequences of '+'-s represent that several papers concluded in the same superior position. One can identify that MSE, CE and $E_{Exp}$ incorporating them are the most often occurring superior measures according to training stability/robustness.}
				\label{table_stability_robustness}
			\end{table}
			
		\subsection{Summary of superiorities}

	    The Tables \ref{table_accuracy}, \ref{table_speed}, \ref{table_stability_robustness} presented the number of researches in which a given measure has some advantageous feature. The Table \ref{table_summary} summarizes the results from the previous tables. It seems that MSE and CE are highly researched measures and have favourable properties such as $E_{ExpAbs}$ which integrates them.
			\begin{table}[ht!]
				\centering

				\begin{tabular}{|c||c|c|c|c|c|c|c|c|c|}
					\hline
					& MSE & SE & CE & CorrE & RE(2) & MI & J & JS & $E_{Exp}$ \\
					\hline \hline
					Accuracy & 8 & 4 & 8 & 4 & 2 & 1 & 2 & 2 & 1 \\ \hline
					Training speed & 7 & - & 1 & 2 & 1 & 1 & 2 & 2 & 4 \\ \hline
					Stability & 1 & 2 & 1 & 2 & 1 & 1 & 1 & 1 & 3 \\ \hline \hline
					Summary & 16 & 6 & 10 & 8 & 4 & 3 & 5 & 5 & 8 \\\hline
				\end{tabular}
				\caption{Summarizing of superiorities from Tables \ref{table_accuracy}, \ref{table_speed},\ref{table_stability_robustness}.}
				\label{table_summary}
			\end{table}
	
	This paragraph structured the review of the actual scientific literature of the previous one and mirrors the Accuracy of the resulted model, training Speed and Stability/Robustness are the most frequent aspects in the comparison of the performances of various error measures. Moreover, \textit{there is (naturally(?)) no best measure among them in none of the viewpoints}, so it is worth to mention a future research topic: the composition of a super-measure is also an exciting and challenging future research field. Measures' positions depend on many factors like distribution and characteristics of noise in the training dataset, outliers, features of their derivatives, etc. \textit{Mean Squared Error (MSE) and Cross-Entropy (CE) have superiorities most frequently, this is the reason why the current paper concentrates on these two measures, especially through the $E_{Exp}$ measure integrating them}.

	\section{The Exponential Error ($E_{Exp}$)} \label{Eexp}
	
	    The key preliminary and excellent scientific result is the introduction of the $E_{Exp}$ measure by \cite{silva2008data}. Silva et al. defined a generally novel measure that integrates MSE, as a statistical measure and two others, CE and ZEDM, as information theory measures. The researchers already performed  similar research before for model training with combined measures \cite{moller_efficient_1993, silva2005neural}. The $E_{Exp}$ is an excellent "marriage" of two different worlds (statistics and information theory), it serves with a static integration of the measures during neural network training. The current manuscript generalizes this measure for introducing a dynamic, self-adaptive and integrated solution, but at first the root scientific result - $E_{Exp}$ - has to be introduced and characterized precisely in the current section.
	
		\subsection{Notations}
		
		The following notations will be used in the description of the mathematical background.
		
		\begin{table}[H]
			\begin{tabular}{@{}ccl@{}}
			\multicolumn{2}{l}{Notations} & \\ 
			\toprule[1pt]
			$ \mathbf{o}^l $ & & The output vector on the $ l $ -th layer \\
			& $l$ & $ 0, \ldots , L $ \\
			& $ L-1 $ & Number of hidden layers \\
			& $ \mathbf{o}^0 $ & Input vector of the MLP \\
			& $ \mathbf{o}^L $ & The output vector of the MLP \\
			\midrule
			$ o_{n_l}^l $ & & The output of the $ n_l $-th output \\
			& & node on the $ l $ -th layer \\
			& $ n_l $ & $ 0, \ldots , N_l $ \\
			& $ N_l $ & The number of output nodes \\
			& & on the $ l $ -th layer \\
			\midrule
			$ w_{n_{l-1}, n_l}^l $ & & Weight from the node $ o_{n_{l-1}}^{l-1} $ to  $ o_{n_l}^l $ \\
			& $ l $ & $ 1, \ldots , L$  (there are not any input \\
			& & weight to the 0-th layer) \\
			\midrule
			$ p $ & & Index of patterns \\
			& $ p $ & $ 1, \ldots , P $ \\
			& $ P $ & The number of patterns \\
			\midrule
			$ i $ & & Index of iteration \\
			& $ i $ & $1, \ldots , I $ \\
			& $ I $ & The number of the iterations \\
			\midrule
			$ \tau $ & & Real number \\
			$ h $ & & Real number \\
			\midrule
			$ e_{N_L} $ & & $ t_{N_L} - o_{N_L}^L $ \\
			& $ t_{N_L} $ & Target value \\
			& $ o_{N_L}^L $ & Output value on the output layer \\
			\midrule
			$ \sigma $ & & Sigmoid activation function \\
			\bottomrule
			\end{tabular}
			\caption{In the representation of the mathematical framework this notations are used.}
			\label{notations}
		\end{table}
		
		As it was mentioned above, $ E_{Exp} $ is a measure, which integrates some of other error types and their properties (\cite{silva2008data}). In the next paragraphs it will be summarized as a pioneering research result and as the main basis of the research reported in the current paper.
		
		\subsubsection{Mean Squared Error (MSE)}
		Mean Squared Error (MSE) is the most commonly applied error function in the data science algorithms for example in artificial neural network training. It is based on the quadratic difference ($e$) between the target ($t$) and the actual ($o$) output.
			\begin{equation}
			MSE = \frac{1}{N} \sum_{p=1}^{P} \sum_{n_L=1}^{N_L} e_{n_L,p}^2
			\end{equation}
		
		\cite{marquardt1963algorithm} built his algorithms already in sixties for the minimization of the same measure. He used only the sum of the squared difference without the normalization (as an average) by pattern (and by output) number.
		
		\subsubsection{Cross-Entropy (CE)}
		
		Cross-entropy is inherited from the information theory measure Kullback - Leibler divergence (sometimes it is referred as Information Gain) \citep{Cover1991}.
		
			\begin{equation} \label{CE}
			\begin{aligned}
			CE = \sum_{p=1}^{P} \left(  \sum_{n_L=1}^{N_L} \left[- t_{p,n_L} \cdot log\left(o_{p,n_L}^L\right)   \right] \right) 
			\end{aligned}
			\end{equation}
		
		\cite{silva2008data} confirmed with more literature sources that "CE is expected to estimate more accurately small probabilities" than MSE.
		
		\subsubsection{Zero Error Density Measure (ZEDM)}
		Zero error density function is the result of \cite{silva_zedm_2005, silva_zedm_2006}, too. It was inspired on entropy criteria as well.
		
			\begin{equation} \label{ZEDM}
			ZEDM = \sum_{p=1}^{P} \left(  \sum_{n_L=1}^{N_L} \left[ h^2 \cdot exp{\left( - \frac { 1 } { 2 h^2 } \cdot (t_{p,n_L} - o_{p,n_L}^L)^2 \right) } \right] \right)
			\end{equation}
		where $h$ is a smoothing parameter of the Gaussian kernel function. It is important to note that during the training process $E_{ZEDM}$ shall be maximized (MSE and CE are to be minimized).
		
		\subsection{The generalized measure $ E_{Exp} $} \label{E_Exp_tau}
		
		Based on the gradient behaviours of the previous three functions, \cite{silva2008data} created a new error function:
			\begin{equation}
			E_{Exp} = \sum_{p=1}^{P} \tau \exp{ \left(  \frac{ \mathbf{e_p}^2}{ \tau} \right)} = \sum_{p=1}^{P} \tau \exp{ \left( \frac1{ \tau} \sum_{n_L=1}^{N_L} e_{n_L,p}^2 \right)}
			\end{equation}
			
		\textit{They reported that according to the value of the parameter $ \tau $ (tau), $ E_{Exp} $ recovers MSE, CE or ZEDM:
		\begin{itemize}
			\item If $ \tau > 0 $ then CE,
			\item If $ \tau < 0 $ then ZEDM,
			\item If $ \tau \to \infty $ then MSE.
		\end{itemize}}
		
		In the published training experiments, the value of $ \tau $ was chosen through separated tests using a set of various (fixed) real numbers, and then the best performing was the selection for the final neural network model training.
		
		\subsection{Characterization of $ E_{Exp} $} \label{eexpappl}
		In practice, $E_{Exp}$ was used e.g. for research on sources of ozone episodes. This research is very important, e.g. because ozone in the troposphere has negative impact on health. \cite{fontes2014can} constructed and optimized Multilayer Perceptron (MLP) neural network model for binary predicting of ozone episodes. To measure the error during the training process they used Cross-Entropy and Exponential Error $E_{Exp}$. They found optimal values for the model parameters (values of $\tau$ and of neuron weights in the hidden layers).  In this paper only positive $\tau$ values were applied. It has to be mentioned that \textit{$E_{Exp}$ was quite sensitive to the choice for the value of $\tau$.}
		
		\subsection{$E_{Exp}$ behaviour in training algorithms}\label{E_exp_opt}
		
\begin{figure}[H]
			\centering
			\includegraphics[scale=0.3]{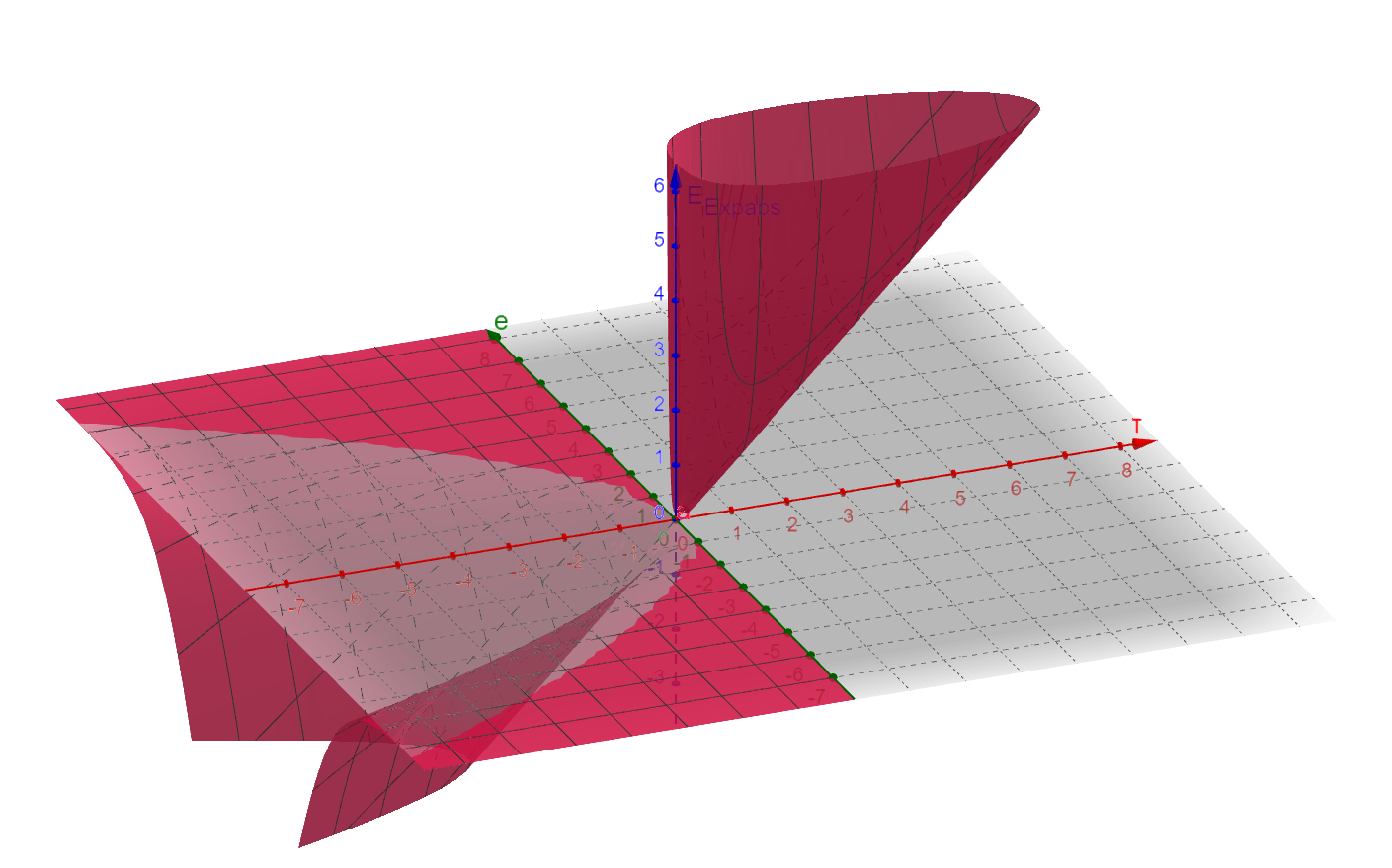}
			\captionsetup{justification=centering}
			\caption{$err_p^{Exp}$ depending on $\tau$ (red axis) and $e$ (green axis) as the difference between target and output}
			\label{fig:geogebraexp}
		\end{figure}

		For one pattern $err_p^{Exp}$ error has the following surface (Figure \ref{fig:geogebraexp}) depending on $\tau$ (red axis) and $e$ (green axis) as the difference between target and output. 
		
		When choosing a fixed value for $\tau$, is cuts from the presented $err_p^{Exp}$ surface,
        \begin{itemize}
			\item if $ \tau > 0 $ a parabola where the minimum point of $err_p^{Exp}$ is accessible only through zero value of $e$ and vice-versa. It means that this measure is completely appropriate for neural network training because it naturally indicates/searches for the minimum target - output difference for all patterns.
			\item If $ \tau < 0 $ then the surface behaves in the same way, the minimization of $err_p^{Exp}$ strives for zero target - output difference.
			\item If $ \tau= 0 $ then $err_p^{Exp}$ goes to infinity, consequently, $\tau$ cannot be zero, moreover when $\tau$ is near to zero the $err_p^{Exp}$ is extremely large, so, in this (small) region this measure is unstable. This instability was mirrored by the performed experiments as well in the preliminary publications.
		\end{itemize}

	    Except $\tau = 0$ (or close to 0), all $\tau$ values are appropriate for neural network training but it has to be kept fixed during the search/training iterations. \textit{However, if the $\tau$ value would have the role to minimize $err_p^{Exp}$ (and $E_{Exp}$ equivalently) it may to go to $\tau \to -\infty$ and also it will indicate $E_{Exp} \to -\infty$. Consequently, the training will never stop and such a solution cannot minimize the difference between target - output values.} Probably, this was the reason why in the various applications of $E_{Exp}$ based neural network training only fixed values for $\tau$ were chosen and a "manual" tuning for its best values were required that is an additional, time consuming calculation/trial as in \cite{silva2008data, silva2014classification, amaral2013using, fontes2014can}.

		%
	
	\section{Measures in training algorithms}
	
	The \hyperref[sec:intro]{Introduction} and the \hyperref[sec:superiority]{Superiority of measures} sections reviewed the characteristics of model error measures but also the applied training algorithm aspect is important as well. There are many various neural network training algorithms, e.g. there is a wide range only of backpropagation (BP) (original, ADAM, stochastic gradient, etc.) calculations. The current paper deals with the Levenberg-Marquardt (LM) algorithm, so, the table \ref{table_training_algorithms} shows the applications of various measures in the backpropagation and also in the LM trainings. It is clearly mirrored that there are various measures that are not yet integrated in LM learning, consequently, there is a scientific niche for these combinations. The current manuscript introduces the extended $E_{exp}$ in the LM algorithm, so, it is a novelty in this training method aspect as well.
	
	    \begin{table}[H]
			\centering
	
			\begin{tabular}{|l||c|c|}
				\hline
				\multirow{2}{40pt}{Applied measure} & 
				BP & LM \\
				& & \\\hline\hline
				MSE & \cite{watrous1992comparison}, \cite{park1995information} &  \cite{marquardt1963algorithm}\\
				& \cite{erdogmus2002error} & \\
				& \cite{silva2005neural, silva2008data, silva2014classification}, \cite{kline2005revisiting} & \\
				& \cite{rimer2006cb3}, \cite{dowson2007mutual} & \\
				& \cite{panin2008mutual}, \cite{rady2011reyni, rady2011shannon} & \\
				& \cite{heravi2018comparison, heravi2018does, heravi2018new, heravi2019new} & \\\hline
				SE & \cite{erdogmus2002error}, \cite{silva2005neural, silva2014classification} & \cellcolor[rgb]{0.557,0.608,0.765} \\
				& \cite{rady2011shannon} & \cellcolor[rgb]{0.557,0.608,0.765} \\\hline
				CE & \cite{watrous1992comparison, park1995information} & \cellcolor[rgb]{0.557,0.608,0.765} \\
				& \cite{silva2005neural, silva2008data, silva2014classification} & \cellcolor[rgb]{0.557,0.608,0.765} \\
				& \cite{kline2005revisiting, rimer2006cb3} & \cellcolor[rgb]{0.557,0.608,0.765} \\
				& \cite{fontes2014can}, \cite{nilsaz2016artificial} & \cellcolor[rgb]{0.557,0.608,0.765} \\
				\hline
				CorrE & \cite{heravi2016new, heravi2018comparison, heravi2018does, heravi2018new, heravi2019new} & \cite{heravi2016new} \\
				& & \cite{heravi2018comparison} \\\hline
				RE & \cite{rady2011reyni}, \cite{silva2014classification} & \cellcolor[rgb]{0.557,0.608,0.765} \\
				\hline
				ZEDM & \cite{silva2008data, silva2014classification} & \cellcolor[rgb]{0.557,0.608,0.765} \\\hline
				$E_{exp}$ & \cite{silva2008data, silva2014classification}, \cite{fontes2014can} & \cellcolor[rgb]{0.004,0.302,0.539} \\\hline
				KL & \cite{park1995information} & \cellcolor[rgb]{0.557,0.608,0.765}\\\hline
				MI & & \cite{thevenaz2000optimization} \\
				& & \cite{dowson2007mutual} \\
				& & \cite{panin2008mutual} \\\hline
				J & \cite{park1995information} & \cellcolor[rgb]{0.557,0.608,0.765} \\
				\hline	
				JS & \cite{park1995information} & \cellcolor[rgb]{0.557,0.608,0.765} \\ \hline
			\end{tabular}
			\caption{Review of the scientific literature on the various applied neural network error measures and on their applications in the backpropagation (BP) and Levenberg-Marquardt (LM) training algorithms.}\label{table_training_algorithms}
		\end{table}
		
	\section{Motivation}
		It is well-known that artificial neural network based modelling is a promising, continuously evolving field of artificial intelligence. It is applicable to solve many kinds of problems in science or life (\cite{viharos_ai_2007}, \cite{viharos_training_2002}). Newer and newer improvements are available in the algorithms considering the scientific research and the applications as well.
		
		It was shown in previous sections that neural network training algorithms can be used efficiently using various types of error measures. Various error measures arise in the literature, MSE and CE are the most popular ones, but there is no "best of" error measure among them for artificial neural network training. The same effect was identified for feature selection where the adaptive and hybrid solution incorporating  statistical and information theory measures resulted a superior algorithm (\cite{viharos_adaptive_2021}, AHFS). The neural network training error measures are typically also inherited from statistics or from information theory, moreover there is a measure ($E_{Exp}$) which integrates three different measures from this two mathematical fields of statistics (a measure: $MSE$) and information theory (two measures: $E_{CE}, E_{ZEDM}$). However, beyond this integrated, excellent measure, \textbf{currently there is no knowledge about a "super, integrated measure" that incorporates almost all of the possible measures, so, there exists a clear challenge in this direction for future scientific research.}
		
		This generalized measure, $E_{Exp}$ was already used in backpropagation algorithms. In these applications the value of $\tau$ was fixed during the training, which means the algorithms used the three measures but with fixed "relative importance". At least two questions may arise: 1, Can the usage of $E_{Exp}$ result in more effective and faster solution for other training algorithms? 2, Is it possible to take advantage of its generalizing property and change the used measure during the training process? Additionally, it is a practical but also a theoretical issue, how to set the fixed value of the (hyper-parameter) $\tau$ for the training?
		
		It is known that LM is a stable and fast ANN training algorithm, according to the state-of-the-art literature MSE is mostly applied in LM algorithms to calculate the error and based on its value the weights of the neural network are modified.
		
		In the current research MSE is replaced by $E_{Exp}$ inside the Levenberg-Marquardt algorithm. In this novel LM method, parameters are optimized so, that during the training also the appropriate value of $\tau$ is changed/searched continuously, because it had been included into the set of training parameters beyond the ANN weights. \textbf{Consequently, a novel type of the original LM algorithm is also introduced.} \textbf{This approach is a dynamic, self-adaptive modification of $\tau$ resulting that at each iteration step the algorithm itself can decide in which direction it is more beneficial to go, namely, dynamic importance can be given to statistics or to information theory training aspects.} It will result in beneficial training behaviour and also in more accurate models than having only one field in focus.
		
		The aim was to develop a stable and fast training solution (as first step, at least the convergence has to be experienced), which performs well on the selected test datasets and \textbf{the proposed, novel algorithm integrates \textit{dynamically} the two big worlds of statistics and information theory that is the key novelty of this paper.}
		
	\section{Definition of the novel measure $E_{ExpAbs}$}
		The section \ref{E_exp_opt} introduced the problem having $E_{Exp}$ as model error measure in case of negative $\tau$. As solution, this paper proposes the novel $E_{ExpAbs}$ having $abs(\tau)$ instead of $\tau$ in its formula. As introduced in paragraph (\ref{E_Exp_tau}), the $\tau < 0$ realizes $E_{ZEDM}$, consequently, the introduced algorithm does not considers this information theory measure during the learning. However, $\tau > 0$ is still realizing $MSE$ (with large $\tau$) and also $E_{CE}$ (with small $\tau$), consequently, the novel algorithm incorporates the parallel consideration of the two worlds of statistics and information theory during training.
		
\begin{figure}[H]
			\centering
			\includegraphics[width=1.0\linewidth]{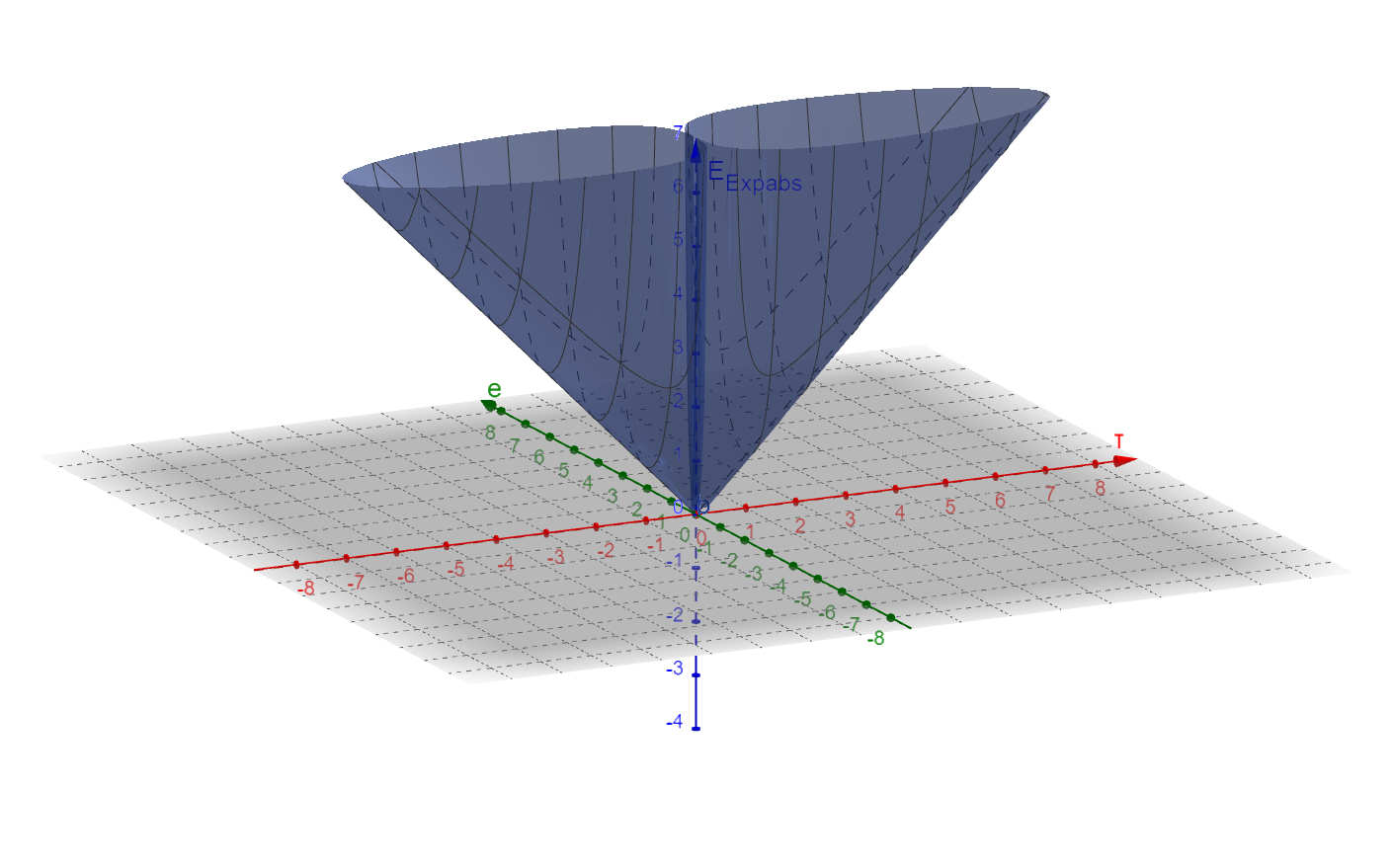}
			\caption{$E_{ExpAbs}$ depending on $\tau$ (red axis) and $e$ as the difference between target and output (green axis)}
			\label{fig:geogebraexpabs}
		\end{figure}

		Based on this suggestion, the following (absolute) error expressions is proposed (Figure \ref{fig:geogebraexpabs}).
		
		For one pattern:
		\begin{equation}
			err_p^{ExpAbs}
			:= |\tau| \exp{ \left( \frac1{|\tau|} \sum_{{n_L}=0}^{N_L} e_{n_L,p}^2 \right)} 
			= |\tau| \exp{ \left( \frac1{|\tau|} \sum_{{n_L}=0}^{N_L} \left( t_{n_L} - o_{n_L}^L \right) ^2 \right) }
		\end{equation}
		
		For P pattern:
		\begin{equation}
			E_{ExpAbs} := \sum_{p=1}^{P} |\tau| \exp{ \left(  \frac{ \mathbf{e_p}^2}{ |\tau|} \right)} = \sum_{p=1}^{P} |\tau| \exp{ \left( \frac1{ |\tau|} \sum_{n_L=1}^{N_L} e_{n_L,p}^2 \right)}
		\end{equation}
		
		So, for one pattern $E_{ExpAbs}$ has the following surface (Figure \ref{fig:geogebraexpabs}) depending on $\tau$ (red axis) and $e$ as the difference between target and output (green axis).
		
	    \textit{The novel $E_{ExpAbs}$ measure was integrated into the Levenberg-Marquard neural network training algorithm and the additional dynamic adaptation of $\tau$ during the learning resulted in the proposed superior algorithm as described in the next paragraphs.}
		
	\section{Introduction of the $E_{ExpAbs}$ measure for the Levenberg-Marquardt algorithm}
		
	Levenberg-Marquard (LM) algorithm is a method for minimization of a non-linear function, which achieves the synthetization of advantages of other optimization algorithms (such as Steepest Descent, Newton, Gauss-Newton). So, it can be said, LM is a fast and stable algorithm, compared to the others mentioned.
	LM handles as two basic elements the Jacobian matrix and the derivative vector of the applied error measure as describes in the next two subsections.
		
		\subsection{Jacobian for weights and $\tau$ update}
		
		$E_{ExpAbs}$ was applied as measure of the error instead of MSE consequently, in the Jacobian matrix, its derivatives were computed with respect to all weights. \textit{An important, additional trick of this novel algorithm was that the changes of $\tau$ parameter is performed together with weight updates during learning as shown in the novel Jacobian matrix \ref{tau_in_matrix}.} The additional last row of this matrix contains the derivatives of $E_{ExpAbs}$ with respect to $\tau$ for all patterns.
		
		\begin{align} \label{tau_in_matrix}
			J_i =
			\begin{bmatrix}
				\frac{\partial err_{1}^{ExpAbs}}{\partial w_{0,1}^1} &  \frac{\partial err_{2}^{ExpAbs}}{\partial w_{0,1}^1} & \dots & \frac{\partial err_{P}^{ExpAbs}}{\partial w_{0,1}^1} \\[2.5ex]
				\frac{\partial err_{1}^{ExpAbs}}{\partial w_{1,1}^1} &  \frac{\partial err_{2}^{ExpAbs}}{\partial w_{1,1}^1} & \dots & \frac{\partial err_{P}^{ExpAbs}}{\partial w_{1,1}^1} \\[2.5ex]
				\vdots & \vdots & \ddots & \vdots \\[2.5ex]
				\frac{\partial err_{1}^{ExpAbs}}{\partial w_{N_1,1}^1} &  \frac{\partial err_{2}^{ExpAbs}}{\partial w_{N_1,1}^1} & \dots & \frac{\partial err_{P}^{ExpAbs}}{\partial w_{N_1,1}^1} \\[2.5ex]
				\vdots & \vdots & \ddots & \vdots \\[2.5ex]
				\frac{\partial err_{1}^{ExpAbs}}{\partial w_{N_1,N_L}^{N_L}} &  \frac{\partial err_{2}^{ExpAbs}}{\partial w_{N_1,N_L}^{N_L}} & \dots & \frac{\partial err_{P}^{ExpAbs}}{\partial w_{N_1,N_L}^{N_L}} \\[2.5ex]
				\textcolor{red}{\frac{\partial err_{1}^{ExpAbs}}{\partial \tau}} &  \textcolor{red}{\frac{\partial err_{2}^{ExpAbs}}{\partial \tau}} & \textcolor{red}{\dots} & \textcolor{red}{\frac{\partial err_{P}^{ExpAbs}}{\partial \tau}} \\
			\end{bmatrix}_{({\left( \text{number of weights} +1 \right)} \times Pattern number)}
		\end{align}
		
		\subsection{Derivative vector of the novel $E_{ExpAbs}$ measure} \label{E_expabs_derivatives}
		The derivatives of $err_p^{ExpAbs}$ with respect to $\tau$ are:
		\begin{equation}
			\begin{aligned}
				\frac{\partial err_p^{ExpAbs}}{\partial \tau}
				&= sign(\tau) \cdot \exp{\left( \frac1{|\tau|} \sum_{n_L = 0}^{N_L} e_{n_L,p}^2 \right)} \cdot \left( 1 - \frac1{|\tau|} \sum_{n_L = 0}^{N_L} e_{n_L,p}^2 \right)
			\end{aligned}
		\end{equation}
		
		Consequently, derivatives of $err_p^{ExpAbs}$ with respect to the outputs on the neural network output layer ($L$-th):
		\begin{equation}
			\begin{aligned}
				\frac{\partial{err_p^{Exp}}}{\partial{o_{n_L}^L}}
				&= - 2 \cdot \left[ \exp{ \left( \frac1{|\tau|} \sum_{{n_L}=0}^{N_L} e_{n_L,p} ^2 \right) } \right] \cdot e_{n_L,p}
			\end{aligned}
		\end{equation}
		
		Calculation of derivatives of the outputs on the $L$-th layer with respect to outputs on the ($L-1$)-th layer:
		\begin{equation}
			\begin{aligned}
				\frac{ \partial{o_{n_L}^L} }{ \partial{o_{n_{L-1}}^{L-1}}}
				= \sigma{ \left( o_{n_L}^L \right) } \cdot \left( 1 - \sigma{ \left( o_{n_L}^L \right) } \right) \cdot w_{n_{L-1},n_L}^L
			\end{aligned}
		\end{equation}
		
		In general:
		
		\begin{equation}
			\begin{aligned}
				\frac{ \partial{o_{n_l}^l} }{ \partial{o_{n_{l-1}}^{l-1}}}
				= \sigma{ \left( o_{n_l}^l \right) } \cdot \left( 1 - \sigma{ \left( o_{n_l}^l \right) } \right) \cdot w_{n_{l-1},n_l}^l
			\end{aligned}
		\end{equation}
		
		Derivatives of of the outputs on the $l$-th layer with respect to a weight to this output calculated as:
		\begin{equation}
			\begin{aligned}
				\frac{ \partial{o_{n_l}^l} }{ \partial{w_{n_{l-1},n_l}^{l}}}
				=  \sigma{ \left( o_{n_l}^l \right) } \cdot \left( 1 - \sigma{ \left( o_{n_l}^l \right) } \right) \cdot o_{n_{l-1}}^{l-1}
			\end{aligned}
		\end{equation}
		
		With these novel calculation methods the proposed, novel, basic LM algorithm is defined exactly.
		
		\section{Testing conditions of the novel, proposed algorithm}
		Having defined each step of the novel algorithm, its comprehensive testing has to be performed as described in the next paragraphs.
		The new algorithm was implemented in Python \cite{python} used Google Colab \cite{colab}. In the testing runs partly Google Colab and the authors institute's, dedicated test server computer was applied (see the Acknowledgement).
		
		\subsection{Methods for faster and more stable algorithm}
		
		The simple combination of Levenberg-Marquardt and $E_{ExpAbs}$ resulted a novel algorithm, however, to further speed up the calculation two additional, \textit{well-know techniques, 'Momentum' \cite{rumelhart_learning_1986} together with the solutions proposed by \cite{tollenaere_supersab_1990} in its algorithm 'SuperSAB' were combined and added to the original LM method, consequently, the paper proposes a further novel training algorithm in this respect as well}.
		
		\subsubsection{Momentum}
		
\begin{figure}[H]
			\centering
			\includegraphics[width=0.8\linewidth]{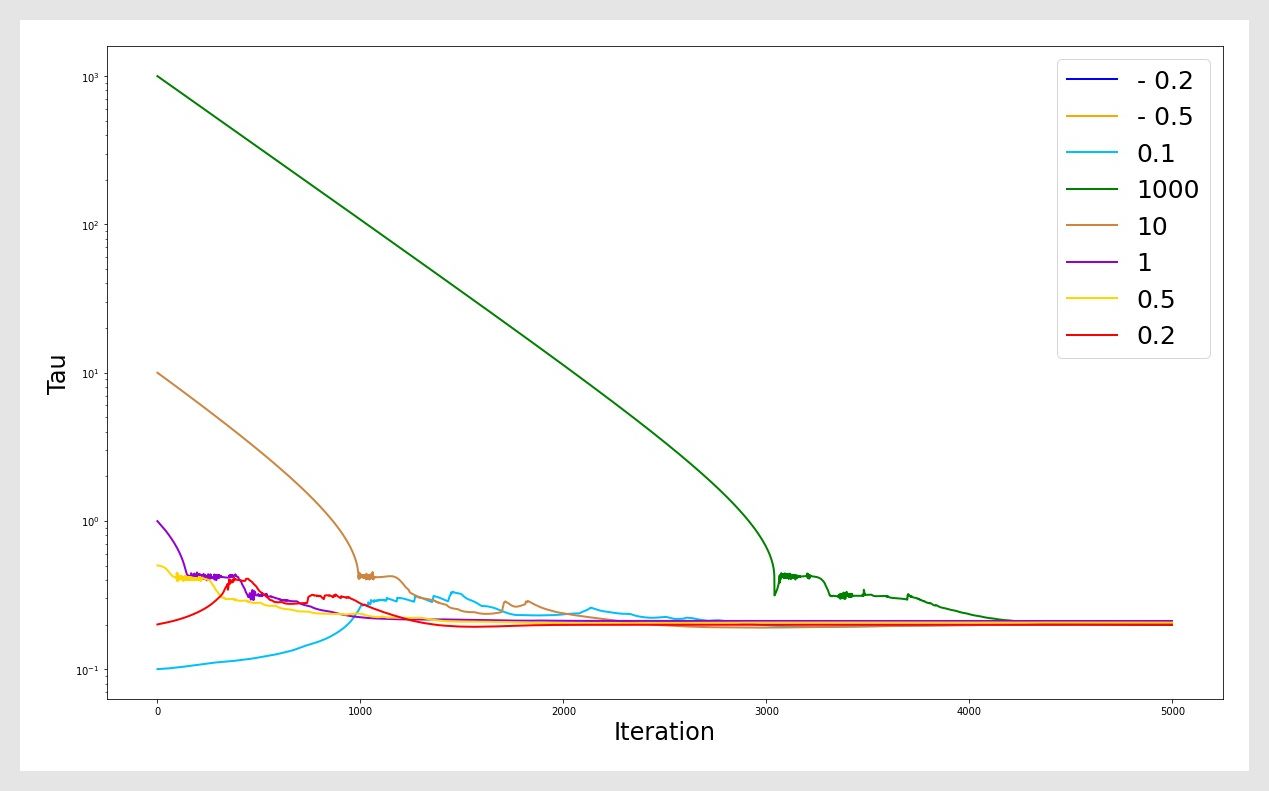}
			\caption{The progress of $\tau$ values from different initial values during neural network training. On this Figure the vertical axis is scaled as logarithmic.}
			\label{fig:allog}
		\end{figure}

		The Figure \ref{fig:allog} highlights the typical resulted character of the $\tau$ curves during training with the novel algorithm on the same dataset. It represents that independently from its initial value the final values of $\tau$-s are (almost) the same at the end of the training processes on the same dataset (e.g., around 0.2 on the Iris dataset).
		
		In the picture \ref{fig:allog} it is shown that the long logarithmic (linear on this logarithmic scale) sections are given at the beginning of tau optimization, and it is longer in case of higher $\tau$-s. This is a typical case for introducing the 'Momentum' extension \cite{rumelhart_learning_1986} as:
		
		\begin{equation} \label{tau_momentum}
		\tau_{k+1} = \tau_k + \alpha \Delta \tau^{previous}
		\end{equation}
	
		Based on these experiences, the expression \ref{tau_momentum} was applied for (only) the $\tau$ parameter, consequently, the newly defined algorithm got a new hyperparameter parameter $\alpha$.
		
		\subsubsection{'SuperSAB'}

		'SuperSAB' is an already traditional neural network speed-up technique that can be used by any grandient based technique. The core idea consists of three components as it was introduced by \cite{tollenaere_supersab_1990}:
		\begin{itemize}
			\item Ordering a parameter called $\eta$ as a multiplier to the derivative of the model weights, consequently, the magnitude of a learning step depends on the product of the multiplier and the derivative itself.
			\item Individual $\eta$ is applied at \textit{all} weights. It means that each training parameter has its own multiplier.
			\item The value of this novel multiplier $\eta$ is adapted dynamically based on the relation between the direction of the preceding learning step and the direction of the derivative. If these directions are the same (meaning that the training performs steps continuously in the same direction) the multiplier $\eta$ is increased slightly but when these directions became opposite (meaning that training changes its direction) then the multiplier $\eta$ is reduced significantly. The small increase but significant decrease relation is important in this speed-up technique. So, when the training realizes a weight update continuously in the same direction (continuous weight increasing or decreasing) the multiplication with $\eta$ increases the step size that was calculated on the basis of the pure derivative. When directions of the weight updates are fluctuating then by the decrease of the multiplier $\eta$ the training performs small, careful steps near the actual learning point.
		\end{itemize}
		
		Details of this speed-up trick is described in \cite{tollenaere_supersab_1990}: as main effect, it can increase the training speed with at least one or two magnitudes than the pure 'Momentum' solution.
		
		\subsubsection{Novel combination of 'Momentum' and 'SuperSAB' in the Levenberg-Marquardt training}
		
		Considering these two speed-update techniques, it is not so simple to include them in the Levenberg-Marquardt (LM) training because the LM training algorithm itself is already a speed-up solution (it integrates the gradient and the Gauss-Newton methods dynamically) and there is a significant risk that the integrated solutions extinguishes each-others' speed-up effects.
		
		As mentioned before, the 'Momentum' speed-up is applied to the $\tau$ parameter and in the same time the 'SuperSAB' method as well. Of course, these techniques have positive effects not only to 'speed-up' but for convergence, stability, etc., but for simple reference only 'speed-up' is mentioned when applying these additional techniques for $\tau$ modifications.
		
		In the final algorithm the following order of the techniques are applied:
		\begin{itemize}
			\item Levenberg-Marquardt (LM) for all network weights and for $\tau$ as well.
			\item 'SuperSAB' speed-up only for $\tau$.
			\item 'Momentum' speed-up for $\tau$ only.
		\end{itemize}
		
\begin{figure}[H]
			\centering
			\includegraphics[scale = 0.2]{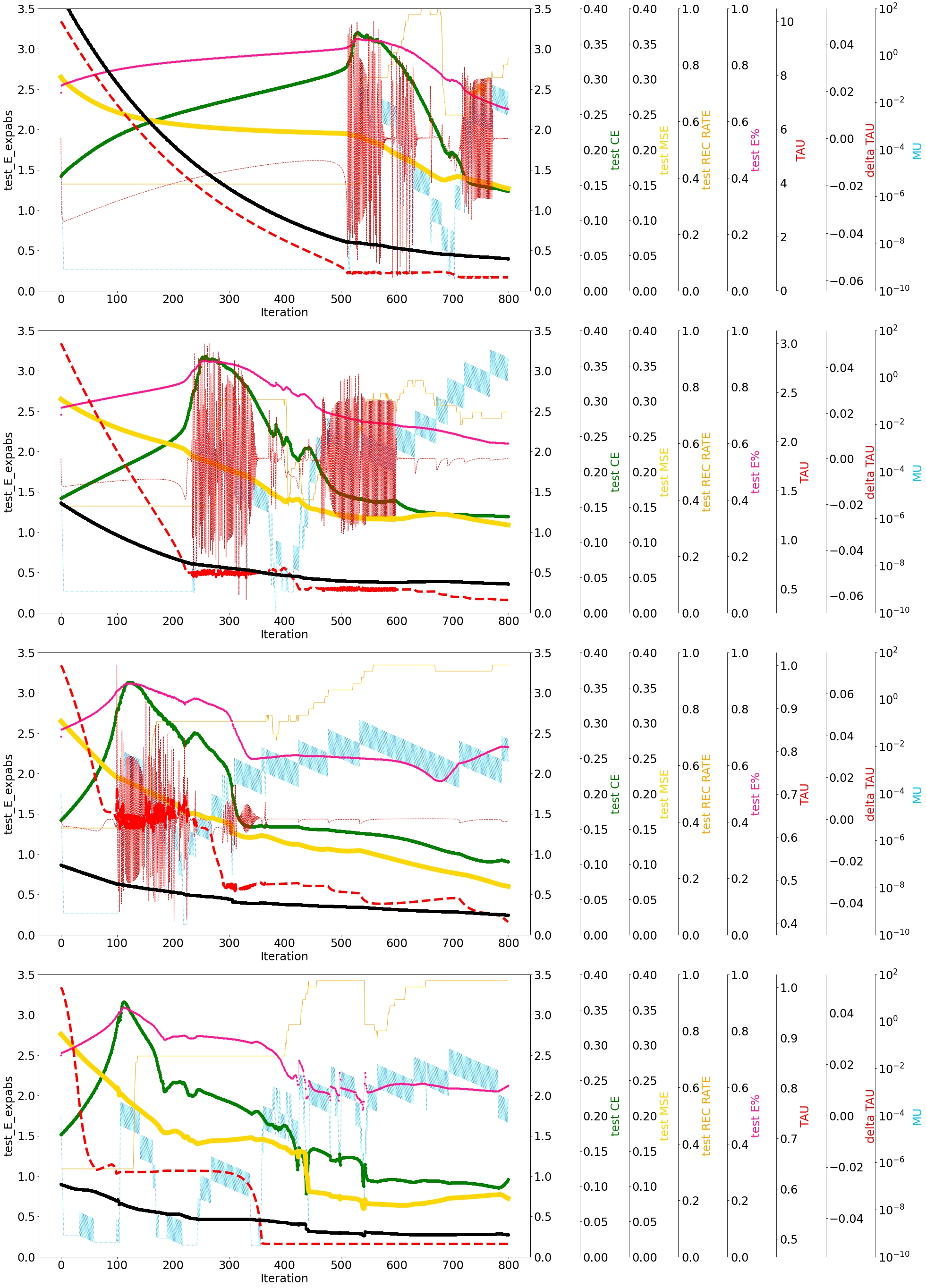}
			\caption{The first three graphs represent the training process on the same network structure with the 'Momentum' method with different initial $\tau$ values (10, 3, 1). The last graph at the bottom mirrors the algorithm accelerated by 'SuperSAB' method with the same initial $\tau 1$ as at the last 'Momentum' (1).}
			\label{fig:mometa}
		\end{figure}

		Figure \ref{fig:mometa} present the different speed-up techniques on the same network structure with different initial $\tau$ parameters. It can be seen that the different training cases resulted in similar training curve characteristics but significantly shorter (quicker) in case of smaller initial $\tau$. However, the method with combined techniques (LM and 'Momentum' and 'SuperSAB') resulted more stable and much quicker training process, with smaller oscillation than the versions with 'Momentum' only.
		
		Other order and further combination of these techniques and also their adaptive tuning were also partly analysed, however, their comprehensive testing is outside of the scope of the current paper, moreover, there is a significant, further research potential in this direction as well. The main aim here was to make the described tests easier, more stable and quicker, so, the combined algorithm was applied, as introduced in pseudo code \ref{alg:pseudo}.
		\\


		
		\par
		
		\begin{breakablealgorithm}\label{alg:pseudo}
		
		\caption{Dynamic speed-up LM}
		
        
        \textbf{Input:} network weights, training and test datasets, validation stepsize for early stopping (\cite{morgan_generalization_1989}),  $\mu$ (combination coefficient of the LM algorithm \cite{HaoYu.2011}), $\alpha$ ('Momentum' parameter), initial $\eta$ value (learning rate of 'SuperSAB' \cite{tollenaere_supersab_1990}), initial $\tau$ value, $m_{max}$ (maximum iteration number for $\mu$ tuning in gradient branch '=5') \\
        \indent \textbf{Output:} optimal weights of network, model accuracy measures and data representing the learning progress \\
        \indent \textbf{Data:} input data, targets (vectors)
		\begin{algorithmic}[1]
			\Procedure{}{}
			
			\While{early stopping or maximum iteration}
				\State $m=1$\; \label{alg:start}
				
				\State extended Jacobian ($\mathbf{J}$) \ref{tau_in_matrix} and difference ($\mathbf{e} = \mathbf{t_{N_L} - o_{N_L}^L}$) computation computation\;
				
		        \textcolor{blue}{/*calculate candidate weights and $\tau$ for the next step of the learning according to the LM rule:*/}
				
				\State $\mathbf{w_{k+1}^{\ast}} = \mathbf{w_{k}^{\ast}} - \left(\mathbf{J^T}\mathbf{J} + \mu I\right) \mathbf{J^T} \mathbf{e}$\; \label{alg:grad_iter} \Comment{\textcolor{blue}{$\mathbf{w^{\ast}}$ means weights extended with $\tau$}}

				\textcolor{blue}{/*increasing the learning speed for  $\tau$ by applying:*/} 
				
				\State $\tau_{k+1} = \eta \cdot \left( \tau_{k+1} - \tau_k \right) + \tau_k $ \Comment{\textcolor{blue}{adaptive learning rate($\eta$)}}
				
				\State $\tau_{k+1} = \tau_{k+1} + \alpha \cdot \Delta \tau$ \Comment{\textcolor{blue}{and 'Momentum'($\alpha$) parameters}}
				
				\State calculate candidate $\left(E_{ExpAbs}\right)_{k+1}$ using $\mathbf{w_{k+1}}$ and $\tau_{k+1}$ \;
				
				\textcolor{blue}{/*compare new and old train errors:*/} 
				
				\If{$\left(E_{ExpAbs}\right)_{k+1}$ $\leq$  $\left(E_{ExpAbs}\right)_k$} \Comment{\textcolor{blue}{Gauss-Newton branch}}
					\State $\mu = \mu \div 10$\;
					
					\State $\mathbf{ w_k} =$ $\mathbf{w_{k+1}}$\;
					
					\State $\left(E_{ExpAbs}\right)_k = \left(E_{ExpAbs}\right)_{k+1}$
					
					\If{$\Delta\tau \cdot \left(\tau_{k+1}- \tau_k \right) \leq 0$} \Comment{\textcolor{blue}{$\eta$ update:}}
					    \State $\eta = \eta \cdot \eta_+$\;
					
					    \State $\Delta\tau = \tau_{k+1}-\tau_k$\;
											
					    \State $\tau_k = \tau_{k+1}$\;

					\Else
						\State $\eta = \eta \cdot \eta_-$\;
						\State $\Delta\tau = 0$\;
					\EndIf
					
				
				\ElsIf{ $\left(E_{ExpAbs}\right)_{k+1}$ $>$ $\left(E_{ExpAbs}\right)_k$} \Comment{\textcolor{blue}{Gradient branch}}
					\If{$m \leq m_{max}$}
						\State $\mu = \mu \cdot 10$\;
						
						\State $m = m+1$\;
						

				    \Else
						\State $\mathbf{w_k} = \mathbf{w_{k+1}}$\;
						
						\State $\tau_k = \tau_{k+1}$\;
						
						\State $\left(E_{ExpAbs}\right)_k = \left(E_{ExpAbs}\right)_{k+1}$
						
					\EndIf

				\EndIf
			
			\EndWhile
			\State \textbf{return} $\mathbf{w_k}$
			\EndProcedure

		\end{algorithmic}
		\end{breakablealgorithm}
		
	    
		\subsubsection{Bounds in the algorithm}
		
		In the proposed, novel LM algorithm two, joint stopping conditions were used. Once the maximum iteration number was bounded (typically by 5000 iteration steps in the test runs) and second, early stopping method determined the optimal stopping point of the learning process (the patience iteration number of the early stopping was 200 in the test runs). According to the paper of \cite{raskutti_early_2014} the first application of early stopping came from \cite{strand_theory_1974}, and probably its first application for ANNs were mentioned at first by \cite{morgan_generalization_1989}). According to the experiences, in most of the cases the early stopping terminating condition became active.

	\section{Test conditions}
	
	Benchmark datasets, testing model structures, (hyper)parameters and other setting are described here.
	
		\subsection{Benchmark datasets}
			
			The key features of the applied benchmark datasets are described in \ref{tab:datasets} table.
			
			\begin{landscape}
			\begin{table} 
			
			\begin{tabular}{|c||c|c|c|c|c|c|c|c|}
				\hline
				& Iris & Linnerud & Wine & Diabetes & Boston & Breast cancer & Ecoli & Glass \\
				\hline \hline
				\# Pattern & 150 & 20 & 178 & 442 & 506 & 569 & 327 & 214  \\
				\hline
				\# Input & 4 & 3 & 13 & 10 & 13 & 30 & 5 & 9 \\
				\hline
				Input types & cont (4) & cont (3) & cont (13) & discrete (1) & discrete (1) & cont (4) & cont (5) & cont (9) \\
				& & & & cont (9) & cont (12) & & & \\
				\hline
				Target types & discrete & cont & discrete & cont & cont & cont & discrete & discrete \\
				\hline
				\# Class & 3 & - & 3 & - & - & 2 & 5 & 6 \\
				\hline
				Class & 50, 50, 50 & - & 59, 71, 48 & - & - & 217, 357 & 143, 77, 35 & 70, 76, 17 \\
				distribution & & & & & & & 20, 52 & 13, 9, 29  \\
				\hline
				\# Output & 1 & 3 & 1 & 1 & 1 & 1 & 1 & 1 \\
				\hline
				Problem & classification & multi-output & classification & regression & regression & classification & classification & classification \\
				types & & regression & & & & & & \\
				\hline
				\# Nodes on & 7 & 6 & 14 & 11 & 14 & 15 & 10 & 15 \\
				hidden layer & & & & & & & & \\
				\hline
				Aim of & Classification of & Predict & Classification of & Predict the & Predict median & Based on & Predict & Glass type \\
				dataset & Iris plant based & physiological & wines based on & progression of & value of homes & calculated & protein & classification \\
				& on the data of & variables of & their chemical & disease one & & datas from & localization & based on its \\
				& its flower & people in & analysis & year after & & digital images & sites & chemical \\
				& & fitness club & & baseline & & predict whether a & & composition \\
				& & & & & & cancer cell is &  & \\
				& & & & & & malignant or &  & \\
				& & & & & & bening &  & \\
				\hline
			\end{tabular}
				\caption{Characteristics of different datasets which was used during the tests. Source of dataset is the UCI Repository \cite{uci2019}.}
				\label{tab:datasets}
			\end{table}
			
			\begin{table}
			\begin{tabular}{|c||c|c|c|c|c|c|}
				\hline
				& Ionosphere & Blood & Parkinson & Thyroid & Vowel \\
				\hline \hline
				\# Pattern & 351 & 748 & 195 & 215 & 990 \\
				\hline
				\# Input & 34 & 4 & 21 & 5 & 10 \\
				\hline
				Input types & cont (34) & cont (4) & cont (21) & cont (5) & cont (10) \\
				& & & & & \\
				\hline
				Target types & discrete & discrete & discrete & discrete & discrete \\
				\hline
				\# Class & 2 & 2 & 2 & 3 & 11 \\
				\hline
				Class & 225, 126 & 178, 570 & 147, 48 & 150, 35, 30 & 90, ..., 90 \\
				distribution & & & & & \\
				\hline
				\# Output & 1 & 1 & 1 & 1 & 1 \\
				\hline
				Problem & classification & classification & classification & classification & classification \\
				types & & & & & \\
				\hline
				\# Nodes in & 15 & 6 & 15 & 8 & 15 \\
				hidden layer & & & & & \\
				\hline
				Aim of & Classification & Predict whether & Discriminate & Predict & Recognition \\
				dataset & of free & the person & healthy people & functioning & of vowels \\
				& electrons & donated blood & from those with & of patient's & from \\
				& in the & in March 2007 & parkinson's & thyroid & different\\
				& ionosphere & & disease & & speakers \\
				\hline
			\end{tabular}
			\end{table}
			\end{landscape}
		
		\subsection {Test model setting}
			During the test process, Multilayer Perceptron (MLP) model was used with one hidden layer. It has to be mentioned that the introduced concept works also for other neural netwkrs structures, e.g., for deep learning architectures as well. The number of hidden nodes on this layer was the sum of input and output parameter number but at least 15, however tests for optimal hidden node numbers were not applied. For classification exercises one-hot encoding was applied for the target variables. The training process was observed on random initialized neural network weights.

		\subsection{Test parameters}
			The algorithm was tested for more initial $\tau$ values: $0.05$, $0.1$, $0.2$, $0.5$, $1$, $2$, $3$, $4$, $5$, $10$, $50$, $100$, $300$, $1000$, $5000$, $10000$. A part of this values was investigated in the research of \cite{silva2008data}, but more other values were chosen for proving the superiority of the proposed method. Small values ($0.05, 0.1, 0.2$) represents mostly CE training direction, but values above 100 are considered as almost only MSE. All starting values between this ranges start the training applying a mixed CE-MSE measure.
			The extra parameter $\alpha$ of the 'Momentum' method was chosen for 0.1. 'SuperSAB' speed-up parameters were: $\eta_+ =1.05, \eta_- = 0.5$ similarly as by \cite{tollenaere_supersab_1990}.

        \subsection{A dedicated test case: comparison to the best performing state-of-the-art, combined error measure based training}\label{adult_intro}
		
		The scientific literature for applying multiple, $E_{Exp}$ related error measures are typically rare. Considering the combination of (one of) the best, state-of-the-art solution is found in \cite{amaral2013using}. They tested different cost function combinations in an auto-encoder training. In the pre-training phase the aim was to find a better initial settings of neural network weights than at random, it is realised by a selected error measure to enable this. As next, this pre-trained network was trained further in the fine-tuning phase, using another error measure. In this two phases, 3-3 different cost function were applied: CE, SSE (MSE), $E_{Exp}$ (with fixed $\tau$ value).
		
		The Adult dataset from the UCI machine learning repository can be applied to compare the performance of the proposed algorithm with this previous, unique, state-of-the-art research result. It is feasible because \cite{amaral2013using} applied the same measures in (pre-, and fine tuning) trainings and in their paper the results were documented appropriately.
		
		
		The network structure for adult dataset training had two hidden layers and 2 neurons in each of them. In the referred research, early stopping method was applied for fixed training (5000 datapoint) and validation (1414 datapoint) sets, and the results were tested on 26147 datapoints.
		
		In the current research the look-ahead parameter of the early stopping was 500 (instead of 25 - look-ahead parameter by the paper of \cite{amaral2013using}), eta\_plus and eta\_minus parameters were 1.02 and 0.3, the initial value of parameter $\tau$ was 10 in the dynamic $\tau$ version, the $\alpha$ parameter of the 'Momentum' was $0.1$.
		
		The algorithm of the current manuscript was tested in a different network structure from the presented one of \cite{amaral2013using} because the current one had only one hidden layer with 8 hidden nodes.
		
		The results of this comparison experiments are presented in section \ref{adult_res}.

	\section{Experimental results}
		
		During the experimental analysis, the results were evaluated by different $\tau$ values and by different adaptation methods (fixed and dynamic $\tau$ during training). This second comparison (fixed or dynamic $\tau$) represent the main benefits and novel behaviour of the proposed algorithm. In the case of dynamic $\tau$, the meaning of $\tau$ means the starting value of it, naturally, because already in the first learning step the novel, proposed algorithm with the $E_{ExpAbs}$ measure modifies its value. The final, various accuracy and other performance measures were investigated through boxplots presenting the means and ranges of them, measured on various (typically 30) repeated tests applied on the benchmark datasets described above.	\textit{It has to be mentioned that these benefits are valid for most of the classification datasets. The non-classification problems showed completely different behaviour, in section \ref{nonclass} this differences are detailed.} In the presented analysis, the benefits were so much spectacular that additional statistical tests (e.g., on averages or deviations of the performance measures) were not needed.
		
		The name \textit{fixed $\tau$ algorithm} or \textit{fixed $\tau$ version} is used when the LM training algorithm with $E_{ExpAbs}$ error measure does NOT change the $\tau$ parameter during training (so $\tau$ is a fixed constant, as in all of the previous state-of-the-art publications referred above). \textit{The fixed $\tau$ algorithm} does not (cannot) include speed-up methods. The name \textit{dynamic $\tau$ algorithm} or \textit{dynamic $\tau$ version} represents dynamically changing $\tau$ values with speed-up techniques of 'Momentum' and 'SuperSAB' in the LM algorithm with $E_{ExpAbs}$ error measure.
		
		
		\subsection{Typical training phases}
			
\begin{figure}[H]
				\centering
				\includegraphics[width=1\linewidth]{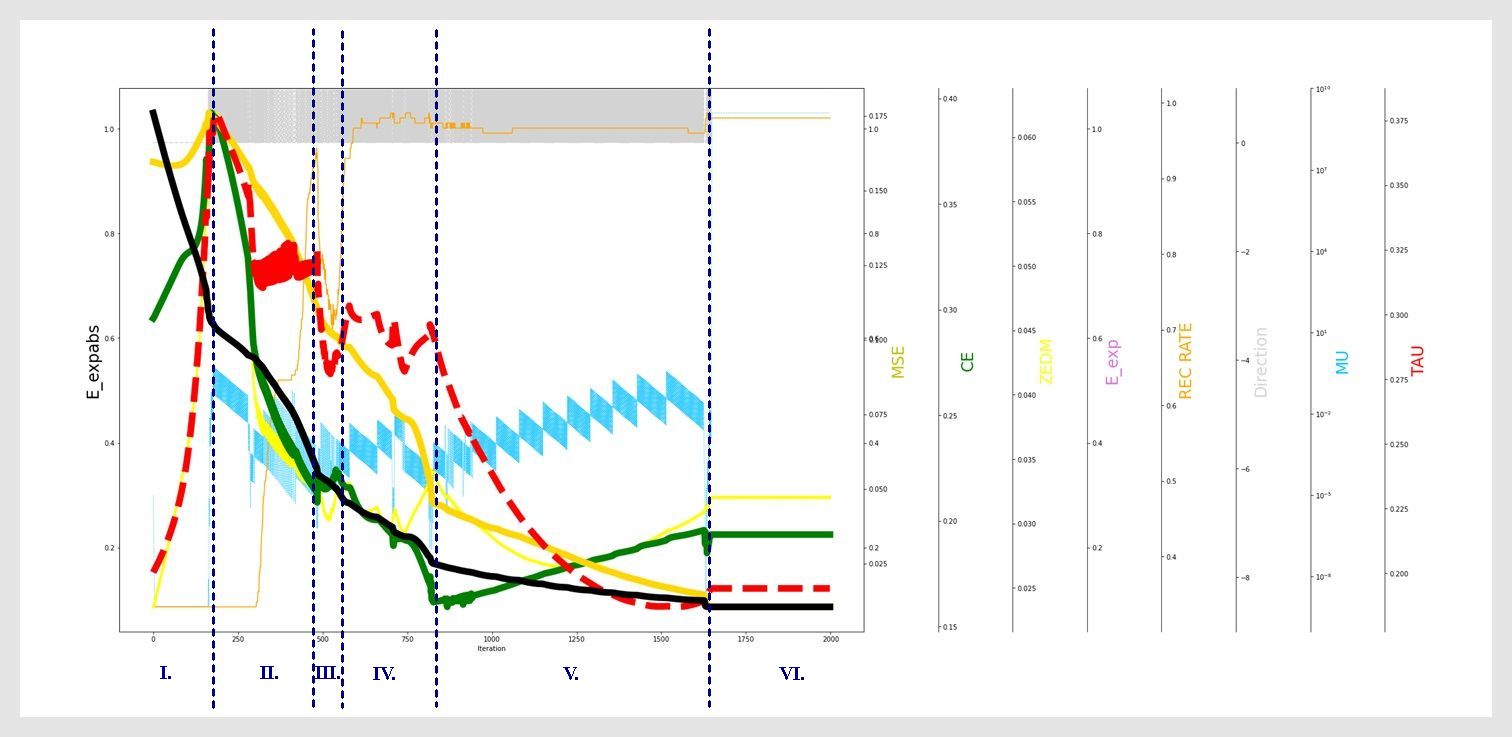}
				\caption{Typical training phases presented on the Iris dataset}
				\label{fig:irisszeletelt}
			\end{figure}

			In this section the typical training phases (marked as roman numerals) are presented on the well-known IRIS dataset (Figure \ref{fig:irisszeletelt}). It is important to mention, that similar phases are observable on the other benchmark sets as well.
			
			The learning phases are differentiated whether the training algorithm is going mainly into the information theory ($CE$) or/and into the classical, statistical ($MSE$) direction.
			
			\begin{enumerate}[label=\Roman*.]
				\item In this phase the values of Mean Squared Error ($MSE$) (bold yellow line) and also Cross-Entropy ($CE$) (bold green line) increase together with the increase of the parameter $\tau$ (dotted, bold red line),  besides that, the target of the training, the value of $E_{ExpAbs}$ (bold black line) decreases. This is a very interesting, starting phase in which both of the statistical and information theory related model performances deteriorates, while the convergence of the overall algorithm is ensured. So, it seems that at this beginning phase the model should deteriorate (set-up) to prepare itself for the next successful learning cycle. Please note, the increase in $\tau$ means that the algorithm prefers going into the $MSE$ (statistics) direction while decrease in $\tau$ represents preference of the $CE$ (information theory) measure. Additionally, the thin yellow line represents the actual recognition rate of the classification while the thin light blue line shows the value of MU ($\mu$) in the Levenberg-Marquardt algorithm of which small value mirrors Gauss-Newton (quick but sometimes unstable) steps but its large number shows gradient learning direction.
				\item Both of $MSE$ and $CE$ are decreasing, while $\tau$ oscillates. In this phase both of the statistical and information theory related model performance improves. The oscillation of $\tau$ mirrors that the algorithm is in an active, explorative and highly convergent phase (in both aspects).
				\item $CE$ increases, $MSE$ comes down, so, the minimised error function ($E_{ExpAbs}$) of the algorithm prefers rather $MSE$ then $CE$. In this phase the $\tau$ starts to strengthen the $CE$ but after a short and intensive decrease in its value it starts to prefer $MSE$.
				\item Both $MSE$ and $CE$ decrease, $\tau$ oscillates through ca. 200 iteration steps. This oscillation seems to be really similar to phase II. resulting in an active, explorative and highly convergent training process. In this phase IV., similar to the phase II., in average, the $\mu$ value of the Levenberg-Marquardt learning algorithm is decreasing which means that the algorithm is moving mainly into the Gauss-Newton training, so, into the quicker convergence direction.
				\item This is again a very interesting part of the learning process because in this phase $MSE$ and $CE$ move to opposite directions. $\tau$ is decreasing continuously and significantly so it forces that the prefered error function is more $CE$ than $MSE$, even if the value of $CE$ is increasing. It can be observed that $\tau$ catches its minimal value in this phase for the complete training process.
				\item All error and training parameter values reaches a constant, fixed value. The direction of algorithm is only gradient (the MU value is relatively high), it is typical in the final training phase using the Levenberg-Marquardt algorithm. $E_{ExpAbs}$ do not decrease any more, the algorithm has reached its minimum point.
			\end{enumerate}

		In the presented research, it seems that the best results of fixed $\tau$ version behaves similarly to \cite{silva2008data} tests, however, the statistical analysis of the introduced approach highlights the new expored perspectives.
		
		\subsection{Comparison of the proposed algorithm according to model accuracy}

		For smaller and also for larger initial $\tau$ values, the dynamic algorithm reached better results, than the fixed valued algorithm. The novel algorithm was compared to the original one based on variety of model training performance measures as $MSE, CE, Recognition Rate (RecRate), E_{ExpAbs} 
		$. First the benefits at the small $\tau$ ranges are presented in the next figures followed by the same representation in the high $\tau$ ranges. Please note that the scale on the x (horizontal) axis of the related next figures are not linear for (initial) $\tau$ but discrete, because the conspicuous differences can be well represent on this way.
		
			\subsubsection{Average model accuracy values of different error measures in the small $\tau$ range}
\begin{figure}[H]
				\centering
				\includegraphics[scale=0.4]{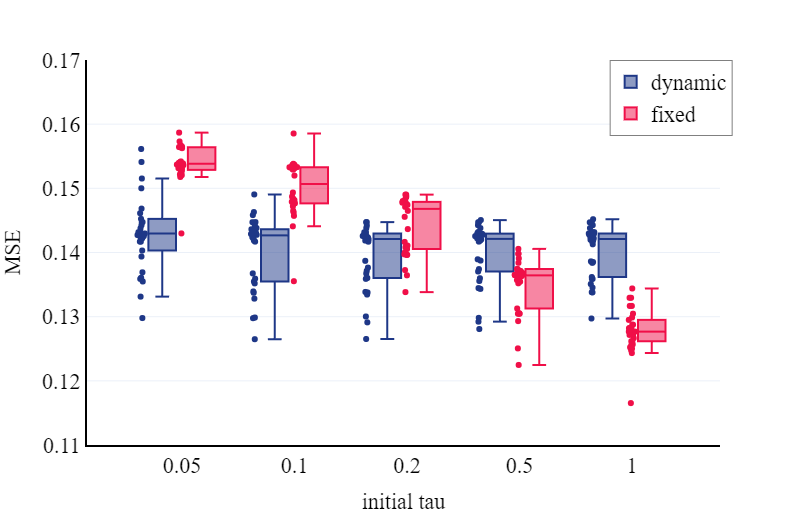}
				\caption{For small $\tau$ values the dynamic algorithm (blue) resulted much smaller median MSE values than then the \textit{fixed $\tau$ version} (red), so, the performance of \textit{dynamic $\tau$ algorithm} is better than the \textit{fixed $\tau$ version}. This figure presents the results on the Blood dataset, however, it is important that other tested classification problem show the same behaviour.}
				\label{fig:blood_smallMSE_avgbetter}
			\end{figure}

			\textit{MSE:}
			The figure \ref{fig:blood_smallMSE_avgbetter} shows that the median of final MSE values are much smaller in case of the proposed, dynamic algorithm for small $\tau$ values (small $\tau$ means much high importance of CE in the training algorithm than MSE). This means that the \textit{new, proposed dynamic $\tau$ algorithm can reach more accurate models than the \textit{fixed $\tau$} based LM, trained by $E_{ExpAbs}$}.
			
\begin{figure}[H]
				\centering
				\includegraphics[scale=0.4]{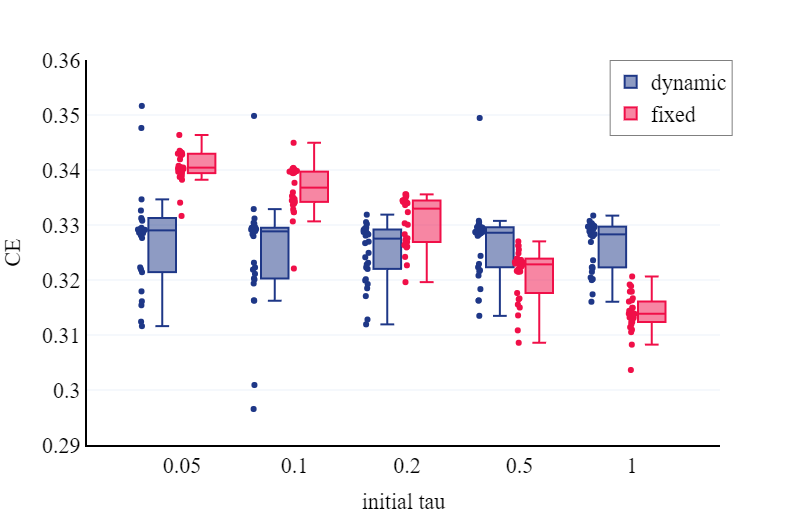}
				\caption{For small $\tau$ values the dynamic speed up algorithm (blue) can reach smaller median CE values than the \textit{fixed $\tau$ version} (red). In case of small $\tau$ values the \textit{dynamic algorithm} can reach better results than the \textit{fixed version} considering CE as evaluation metric. This figure presents the results on the Blood dataset, however, it is important that other tested classification problem show the same behaviour.}
				\label{fig:blood_smallCE_avgbetter}
			\end{figure}

			\textit{CE:}		
			Similar behaviour could be observed considering CE on figure \ref{fig:blood_smallCE_avgbetter}. The median values of dynamic training processes are much lower than the fix version for small $\tau$ values. In the paper of \cite{silva2008data} it was mentioned that $E_{Exp}$ with $\tau > 0$ behaves like CE, however, the actual results show that this limit is not 0 (the limit is larger). Considering the runs with $\tau>0$, the results can be divided into two parts. It seems that, the fixed $\tau$ version algorithm works well between ~0.1 and 100 $\tau$ values, however \textit{the proposed dynamic $\tau$ version can found appropriate solution for all (also smaller) initial $\tau$ values}.
			
\begin{figure}[H]
				\centering
				\includegraphics[scale=0.4]{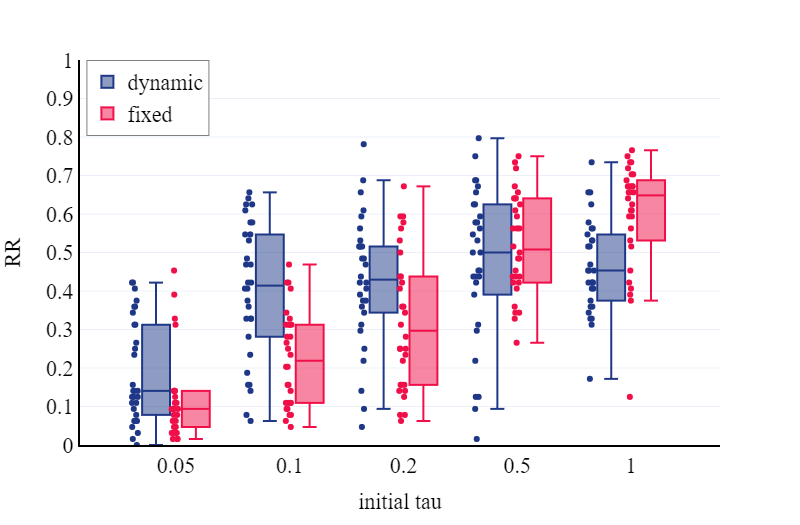}
				\caption{For small $\tau$ values the dynamic algorithm (blue) can reach much higher median Recognition Rate (RR) values than the fixed $\tau$ version (red). Considering RR, \textit{dynamic version} it is able to reach better result than the \textit{fixed version}. This figure presents the results on the Glass dataset, however, it is important that other tested classification problem show the same behaviour.}
				\label{fig:glass_smallRR_avgbetter}
			\end{figure}

			\textit{RecRate:}
			For small $\tau$ values, benefits were identified considering the model Recognition Rate as well. The figure \ref{fig:glass_smallRR_avgbetter} presents the great performance of speeded-up, dynamic modified LM version compared to fixed $\tau$ valued LM. The median values in case of \textit{dynamic $\tau$ version} are much higher than for the \textit{fixed $\tau$ version}, which mirrors the \textit{dynamic algorithm being significantly superior to the fixed version, for small $\tau$ values}.
			
\begin{figure}[H]
				\centering
				\includegraphics[scale=0.3]{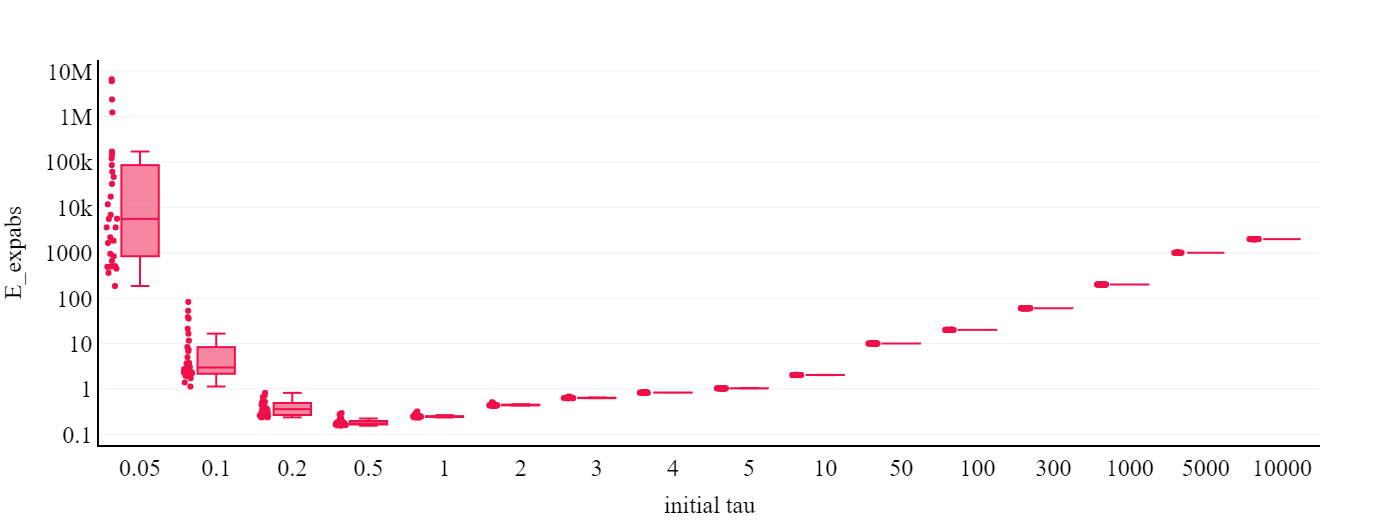}
				\caption{$E_{ExpAbs}$ with logarithmic y axis on Ecoli dataset. It represents that it is useful to normalize $E_{ExpAbs}$ for the comparison. This figure presents the results on the Ecoli dataset, however, it is important that other tested classification problem show the same behaviour.}
				\label{fig:ecoli_allEexpabs_logfix}
			\end{figure}

			\textit{$E_{ExpAbs}$:}
			For \textit{fixed $\tau$ algorithm}, the final $E_{ExpAbs}$ values are (naturally) on a very wide range - presented on figure \ref{fig:ecoli_allEexpabs_logfix} - depending on value of the $\tau$ parameter. To eliminate this problem for the comparability all final $E_{ExpAbs}$ values are presented on logarithmic scale.
		
			In case of small $\tau$, the ranges of 
			final $E_{ExpAbs}$ are slimmer, so, the dynamic algorithm finds similar optimal values for different initial networks than the fix version, \textit{consequently, its more robust in accuracy deviation aspects}.
			\begin{figure}[H]
				\centering
				\includegraphics[scale=0.4]{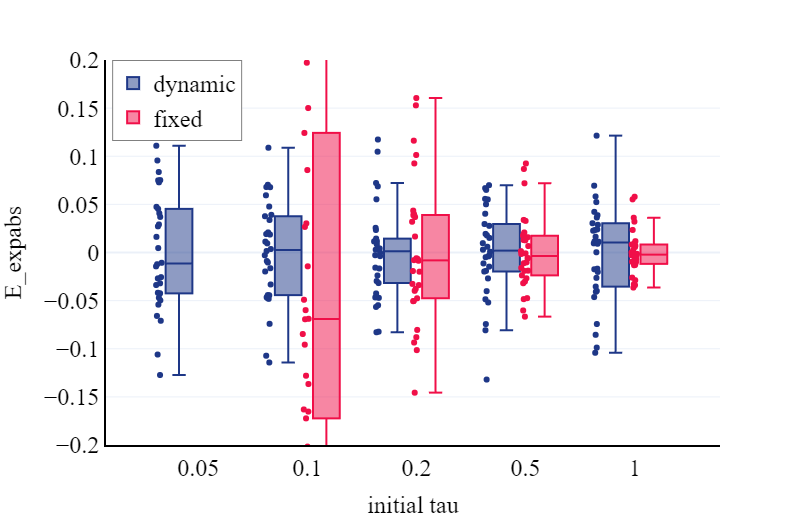}
				\caption{The final $E_{ExpAbs}$ ranges are slimmer in case of dynamic $\tau$ version (blue) than with the fixed $\tau$ version training algorithm. (The points of values are normalized with averages per $\tau$ values.)
				This figure presents the results on the Parkinson dataset, however, it is important that other tested classification problem show the same behaviour.}
				\label{fig:parkinson_smallEexpabs_rangeslimmer}
			\end{figure}
				
			\subsubsection{Average final model accuracy values of different error  measures in the large $\tau$ range}
			
\begin{figure}[H]
				\centering
				\includegraphics[scale=0.33]{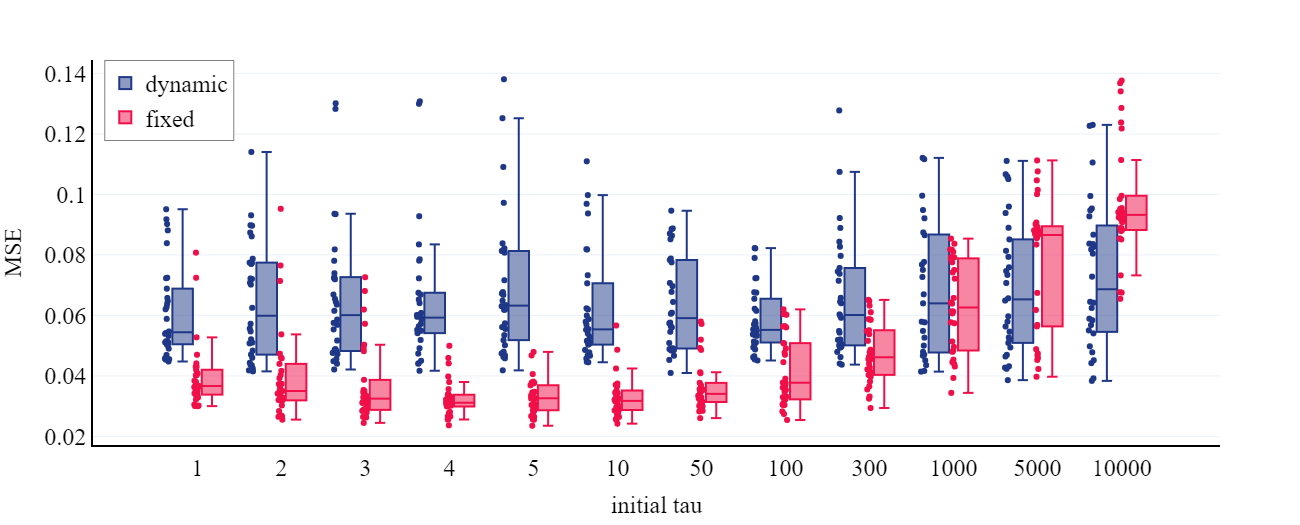}
				\caption{For high $\tau$ values the dynamic algorithm (blue) can reach smaller median MSE values than the fixed $\tau$ version. This figure presents the results on the Ecoli dataset, however, it is important that other tested classification problem show the same behaviour.}
				\label{fig:ecoli_highMSE_acc}
			\end{figure}

			In the picture \ref{fig:ecoli_highMSE_acc} it could be seen that at higher $\tau$ (high $\tau$ means that mainly the MSE is considered during training and the CE has very small, close to no importance) values the \textit{dynamic algorithm can serve with better performance compared to the fixed algorithm}.
			
			\textit{MSE:}			
			There is a range of $\tau$-s where the MSE values are smaller for the fixed version, but at higher $\tau$ values the MSE measures are higher (wronger) for the dynamic algorithm. In the paper of \cite{silva2008data} it was proved that at $\tau \to \infty$, the $E_{Exp(Abs)}$ error measure behaves like the MSE based training. The research results of this manuscript mirror that this is not true above a certain limit. For the $E_{ExpAbs}$ optimized fixed $\tau$ algorithm, the MSE values in case of high $\tau$ does not go according to the statement of \cite{silva2008data},  but the opposite was experienced. This raises the possibility of the existence of an optimal $\tau$ value, moreover/or an optimal $\tau$ range. It is important to note that this value/range is unknown before the training process.
			
			\textit{CE:}			
			For CE, this attitude is less spectacular (Fig. \ref{fig:ecoli_highMSE_acc}), but also observable considering the median values belonging to higher $\tau$ values. The final status of the network reach lower CE values in case of dynamic version, which means it is more accurate in for this high $\tau$ range.
		
			\begin{figure}[H]
				\centering
				\includegraphics[scale=0.33]{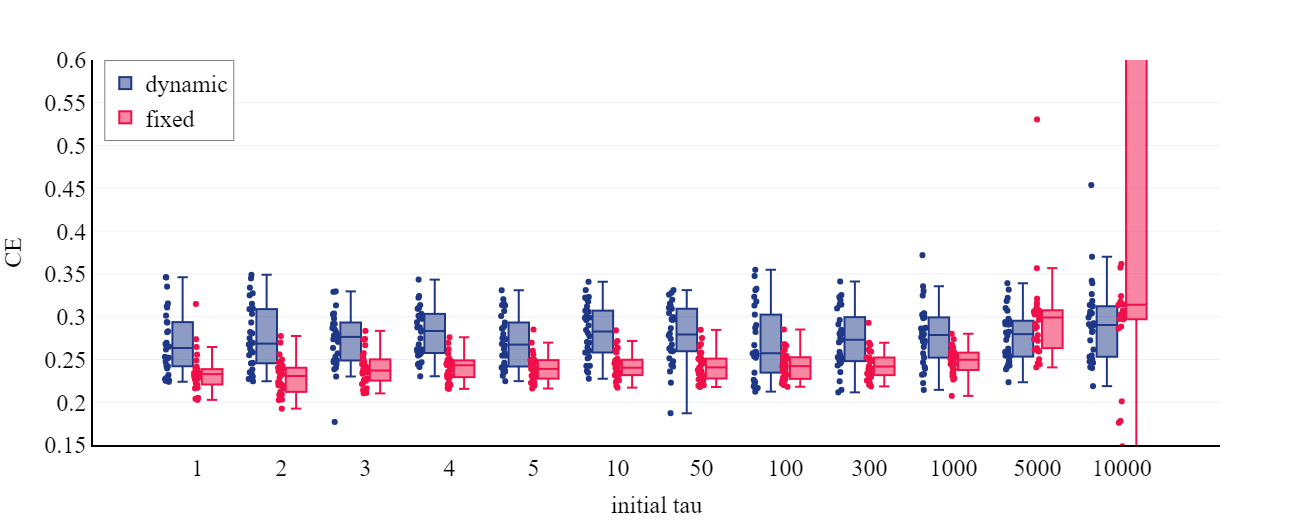}
				\caption{For high $\tau$ values the dynamic algorithm (blue) serves with smaller median CE values than the fixed $\tau$ version. This figure presents the results on the Parkinson dataset, however, it is important that other tested classification problem show the same behaviour.}
				\label{fig:parkinson_highCE_acc}
			\end{figure}
			
\begin{figure}[H]
				\centering
				\includegraphics[scale=0.33]{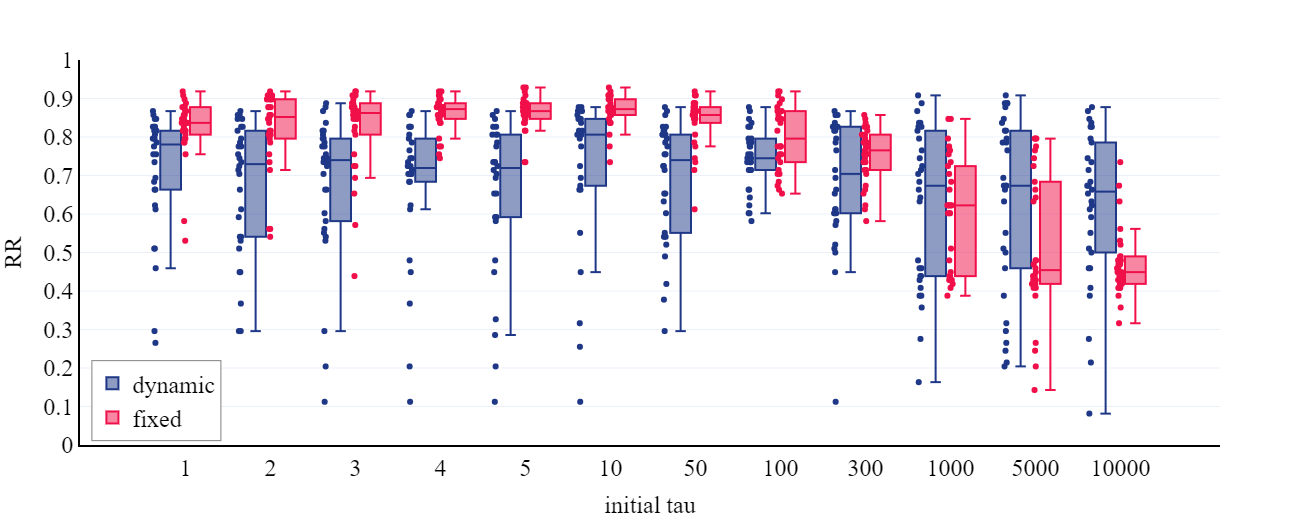}
				\caption{For high $\tau$ values the dynamic speed up algorithm(blue) can reach higher median Recognition Rate values than the fixed $\tau$ version. This figure presents the results on the Ecoli dataset, however, it is important that other tested classification problem show the same behaviour.}
				\label{fig:ecoli_highRR_acc}
			\end{figure}

			\textit{RecRate:}
			The recognition rates of the final models show also this behaviour, considering the Recognition Rate medians belonging to high $\tau$ values (Fig. \ref{fig:ecoli_highRR_acc}).
			
		\subsection{Evaluation and comparison of the proposed algorithm according to training speed}
			
			The success of the training speed improvement and the superiority of the dynamic algorithm over the fixed one was measured.
			
			\subsubsection{Successful speed increase}
			
\begin{figure}[H]
				\centering
				\includegraphics[scale = 0.4]{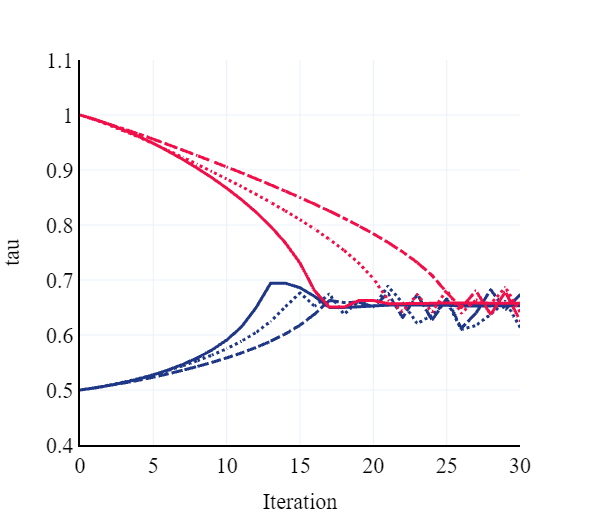}
				\caption{Different speed-up methods for dynamic $\tau$ version in case of the IRIS data set: the solid lines present the speed-up method with 'Momentum' and 'SuperSAB', dotted lines show the algorithm with the 'Momentum' technique, and dashed line draw the simple dynamic algorithm. The $\tau$ values are seen on y axis. The accelerations and the pure dynamic algorithm can find the same, nearly optimal $\tau$ value but with different, slower speeds.}
				\label{fig:iris_smalltau_speedup}
			\end{figure}

\begin{figure}[H]
				\centering
				\includegraphics[scale = 0.4]{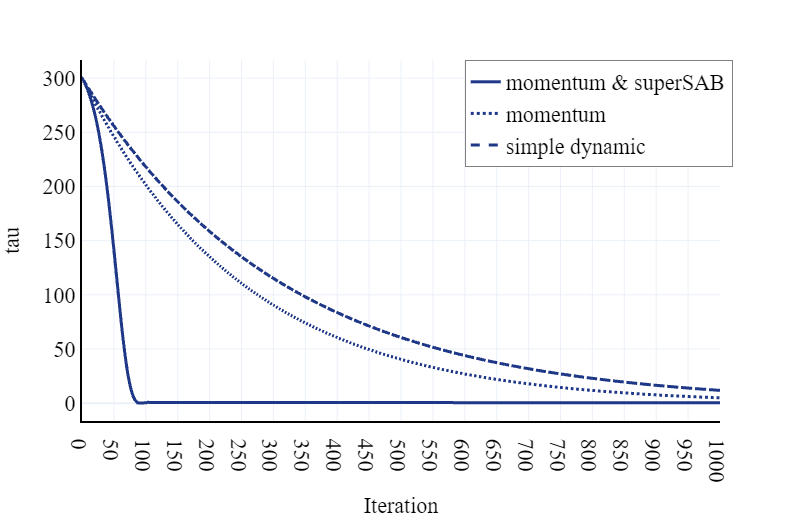}
				\caption{Enormous acceleration are observed for higher $\tau$ values. The best speed-up method found the appropriate $\tau$ value during less then 100 iteration, however the other methods can learn significantly slower.}
				\label{fig:iris_hightau_speedup}
			\end{figure}

\begin{figure}[H]
				\centering
				\includegraphics[scale = 0.25]{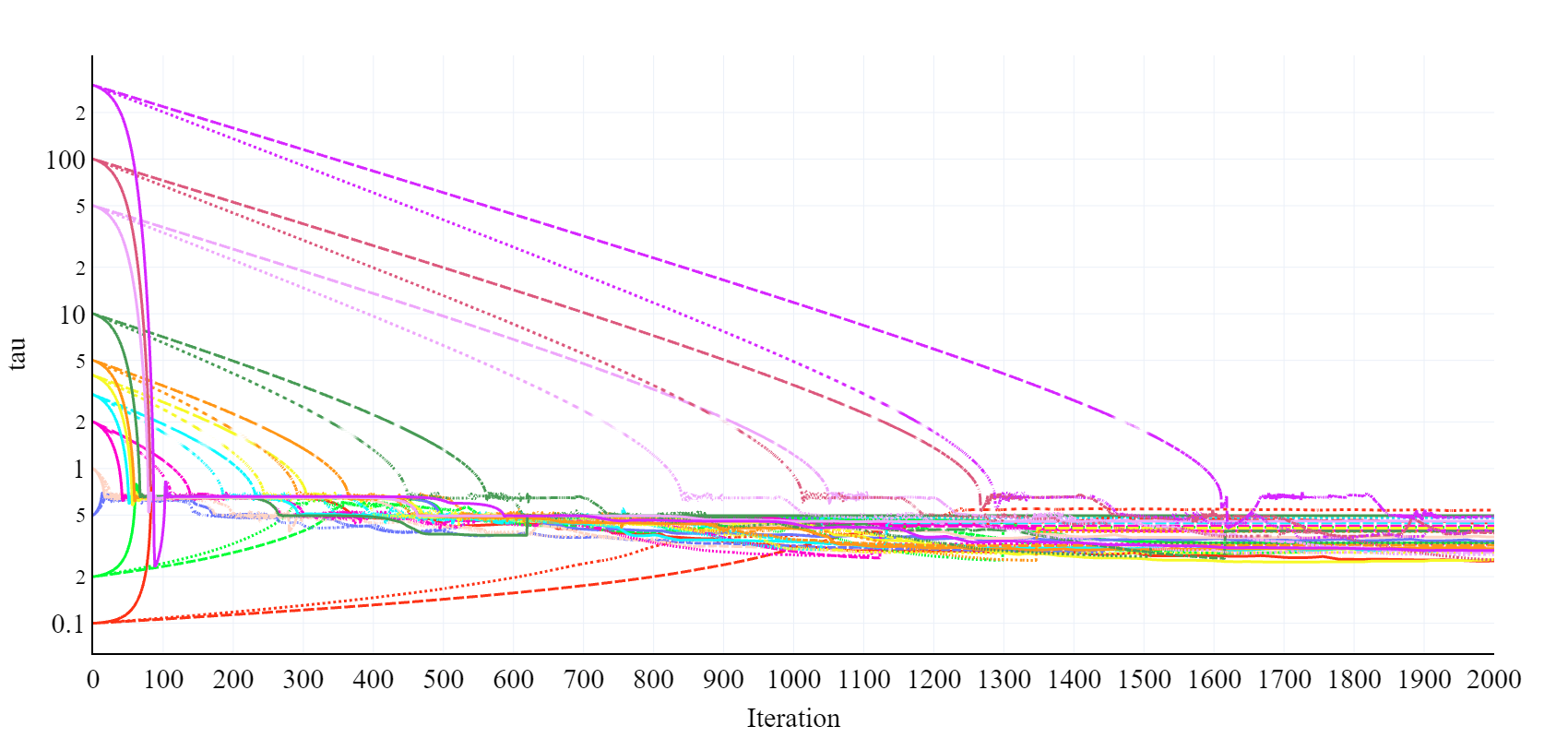}
				\caption{Various training curves of $\tau$ on the IRIS dataset: the solid lines present the speed-up method with 'Momentum' and 'SuperSAB', dotted lines the algorithm with the 'Momentum' technique, and dashed line the simple dynamic algorithm. Please, consider that the Y axis is logarithmic.}
				\label{fig:iris_alltau_speedup}
			\end{figure}

			The figure \ref{fig:iris_smalltau_speedup} shows three different versions of the dynamic $E_{ExpAbs}$ based LM algorithm. When the starting $\tau$ values are close to the final (optimal) one, the speed increase is less, however, the combined application of 'SuperSAB' and 'Momentum' serves with enormous training speed increase when starting far from the (previously unknown) final $\tau$ values (Fig. \ref{fig:iris_hightau_speedup}). This speed increase is consequent in the complete range of starting $\tau$ values (Fig. \ref{fig:iris_alltau_speedup}).
			
		\subsubsection{The novel, dynamic algorithm is a more robust training than the fixed version}
			
\begin{figure}[H]
				\centering
				\includegraphics[scale = 0.55]{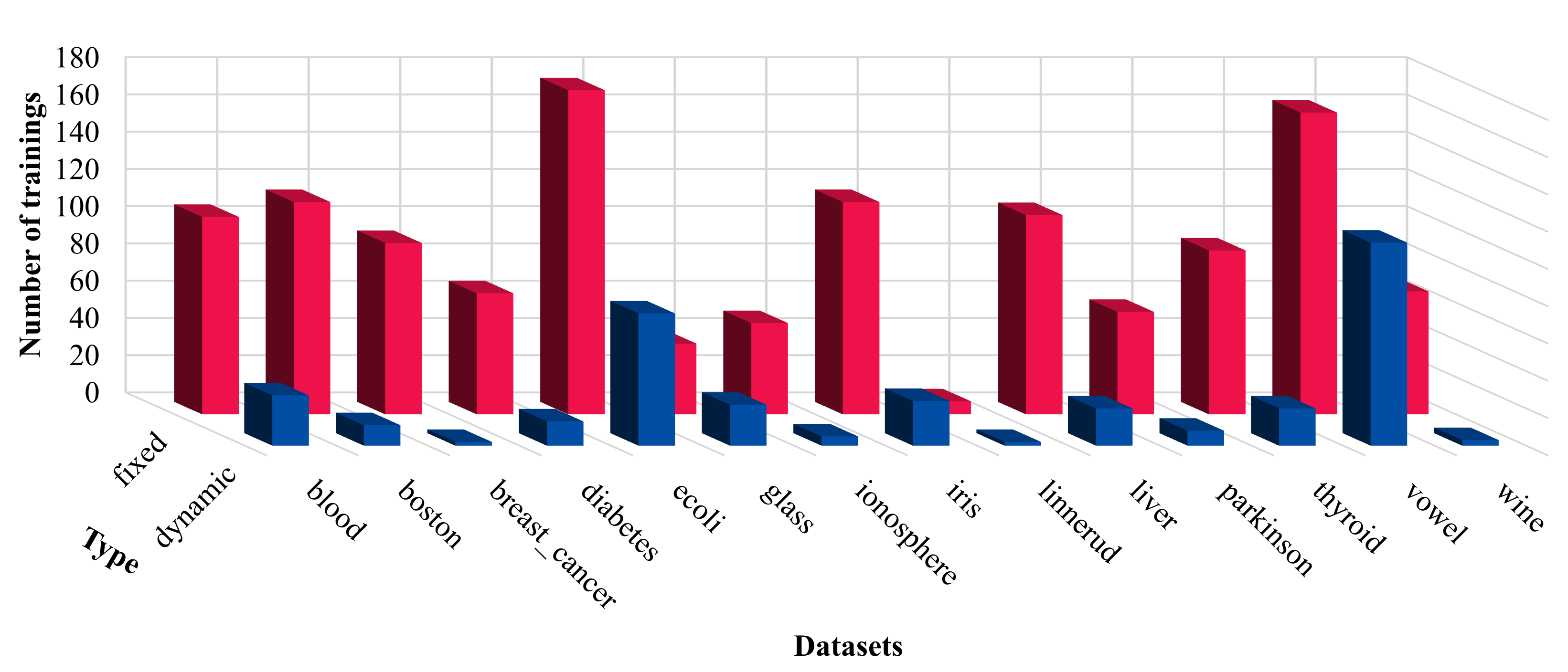}
				\caption{The proposed, novel, dynamic $\tau$ version training is significantly quicker than the fixed version.}
				\label{fig:Iteration_number_over_datasets}
			\end{figure}

			The training process was bounded by a (relatively high) maximum number iteration (5000 iteration steps), but in most of the cases the applied early stopping technique was active (its preselected patience iteration number was 200). During the test experiments only in around 5-6\% of all of the runs were stopped by the maximum iteration number and in far over 90\% ratio was the early stopping active, consequently, these benchmark cases can be trained with much smaller learning steps (than 5000 iteration number). It means that reaching of the iteration step limit (5000) is an indication that the training felt into a local, not optimal minima. 
			The Fig. \ref{fig:Iteration_number_over_datasets} shows how many times the 5000 step number limit was reached using the same number of training's for the fixed and for the novel, dynamic algorithm as well (total training number for one dataset for one version was 480 (= 30 repetitions x 16 different/initial $\tau$)).
			
			Because the training step limit was applied much often (2-10 times more often) by the fixed version than with the novel, proposed, dynamic $\tau$ version algorithm, it is proven that the proposed algorithm falls with much smaller probability to local, not optimal minima, consequently, \textit{the novel algorithm is much more robust considering the training process and its result}.
			
			It has to be mentioned that these additional robustness results are received because the $\tau$ parameter is tuned dynamically during training \textit{that is not possible with the original, fixed $\tau$ version, because there the $\tau$ is constant}.
			
		\subsection{Evaluation and comparison of the proposed algorithm according to training stability/robustness}
		
		The dynamic algorithm is independent of initial $\tau$ evaluated on the different model accuracy measures.
		
			\subsubsection{The average resulted model accuracy measures for different initial $\tau$ values are the same} \label{same_average}
			
\begin{figure}[H]
				\centering
				\includegraphics[scale=0.3]{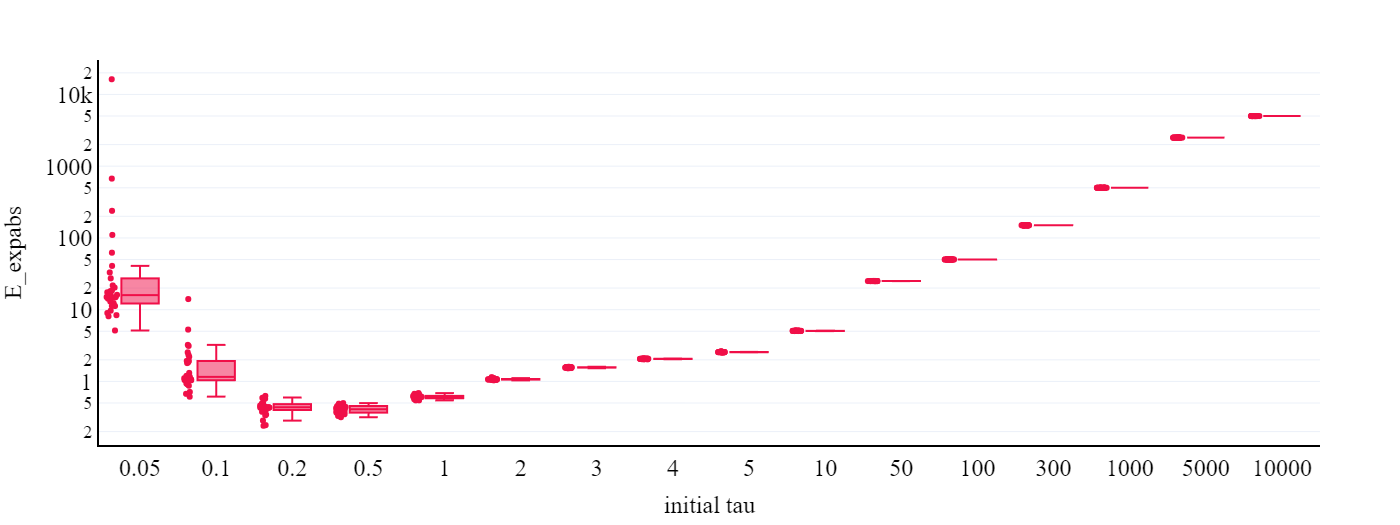}
				\caption{The medians of final $E_{ExpAbs}$ values are different per $\tau$ values. This figure mirrors that the \textit{fixed $\tau$ version} highly depends on the $\tau$ values.This figure presents the results on the Ionosphere dataset, however, it is important that other tested classification problem show the same behaviour.}
				\label{fig:ionosphere_allEexpabs_avgtauind_fix_log}
			\end{figure}

\begin{figure}[H]
				\centering
				\includegraphics[scale=0.3]{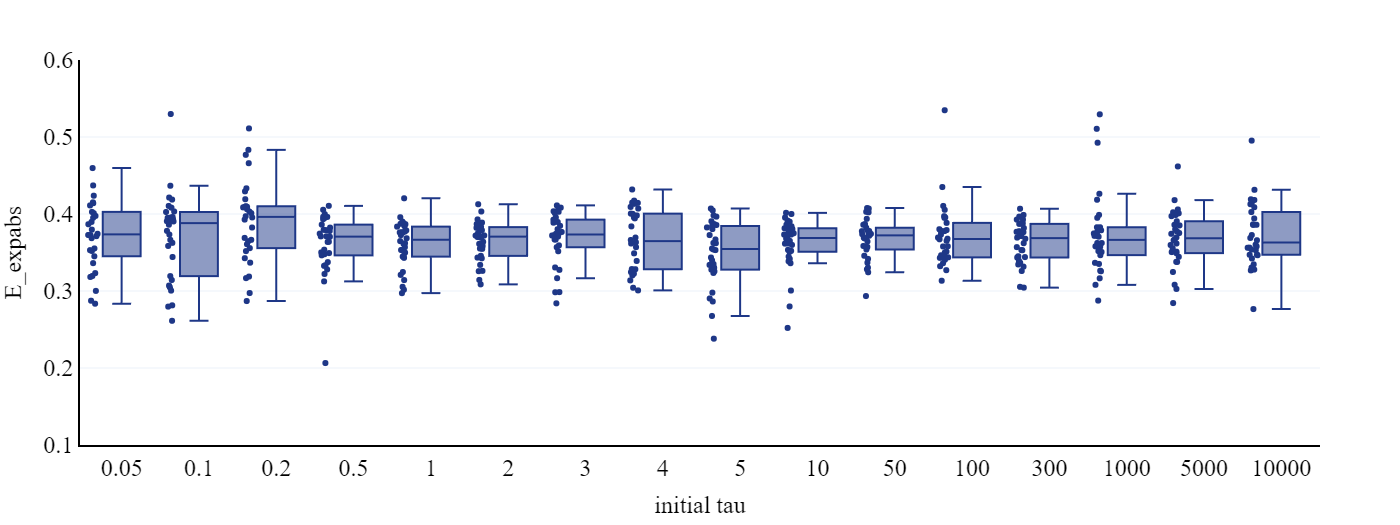}
				\caption{The medians of final $E_{ExpAbs}$ values are similar for all of the tested, initial $\tau$ values, which shows the starting $\tau$ independence of the  dynamic $\tau$ algorithm. This figure presents the results on the Ionosphere dataset, however, it is important that other tested classification problem show the same behaviour.}
				\label{fig:ionosphere_allEexpabs_avgtauind_dyn}
			\end{figure}

			\textit{$E_{ExpAbs}$:}			
			The Fig. \ref{fig:ionosphere_allEexpabs_avgtauind_dyn} mirrors that the final model accuracy measure, $E_{ExpAbs}$ has almost the same value at the end of the training process, consequently, independently from the starting, initial $\tau$ value. This is an important robustness feature of the proposed, novel dynamic training algorithm. (The dependence for fixed algorithm is proofed, because the $\tau$ values do not change during the training process which is presented in Fig. \ref{fig:ionosphere_allEexpabs_avgtauind_fix_log})
			
\begin{figure}[H]
				\centering
				\includegraphics[scale=0.3]{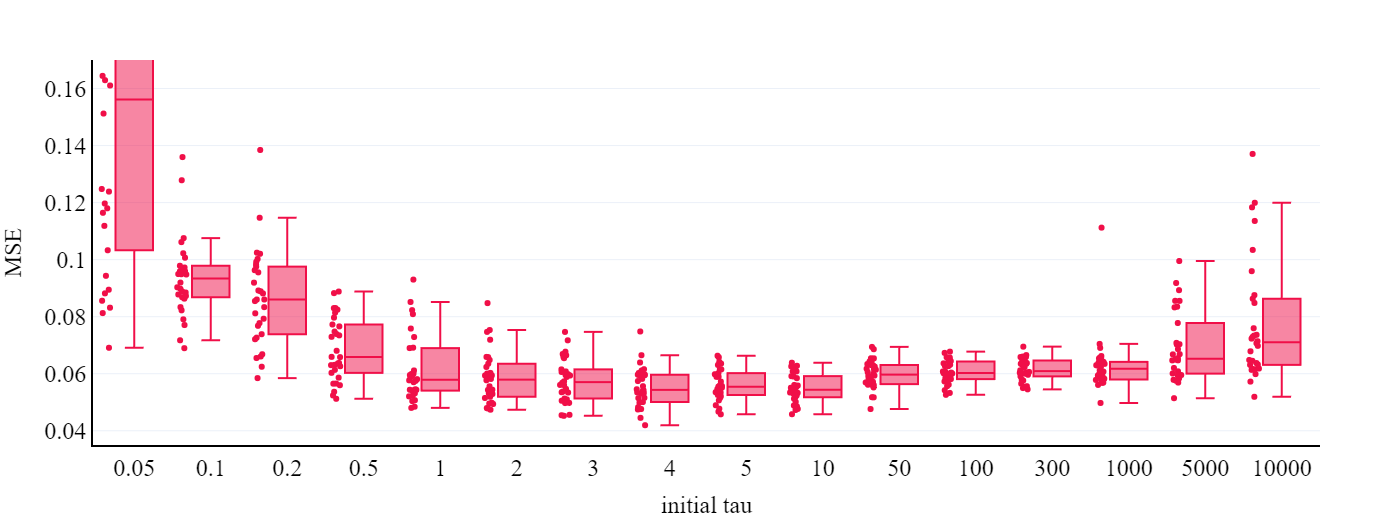}
				\caption{The medians of final MSE values are different per $\tau$ values. This figure mirrors that the \textit{fixed $\tau$ version} highly depends on the $\tau$ values. This figure presents the results on the Glass dataset, however, it is important that other tested classification problem show the same behaviour.}
				\label{fig:glass_allMSE_avgtauind_fix}
 			\end{figure}

\begin{figure}[H]
				\centering
				\includegraphics[scale=0.3]{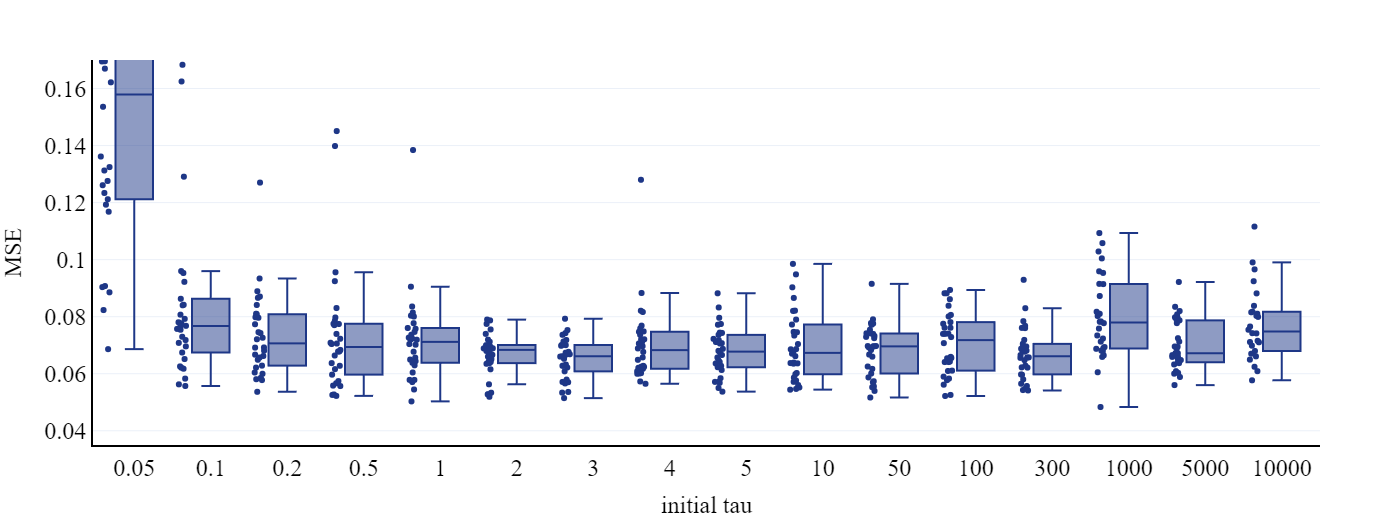}
				\caption{The medians of final MSE values are similar for the tested, inital $\tau$ values, which shows the $\tau$ independence of the dynamic $\tau$ algorithm. This figure presents the results on the Glass dataset, however, it is important that other tested classification problem show the same behaviour. (Only the result at 0.05 initial $\tau$ is an exception but in this case both algorithms are unstable as detailed before)}
				\label{fig:glass_allMSE_avgtauind_dyn}
			\end{figure}

			\textit{MSE:}			
			The comparison of the fix (Fig. \ref{fig:glass_allMSE_avgtauind_fix}) and dynamic (Fig. \ref{fig:glass_allMSE_avgtauind_dyn}) versions for the final model MSE measure shows that the dynamic version serves with similar median values for all initial $\tau$ values but the fixed version is highly sensitive on the selected $\tau$ value. This feature for the fixed version was already extensively analysed by \cite{silva2008data, silva2014classification, fontes2014can, amaral2013using}, the researchers also identified this sensitive behaviour and partly gave a pioneering, two-step approach. However, it will be shown later that the proposed, dynamic version serves with more advanced, more accurate models.
			
\begin{figure}[H]
				\centering
				\includegraphics[scale=0.3]{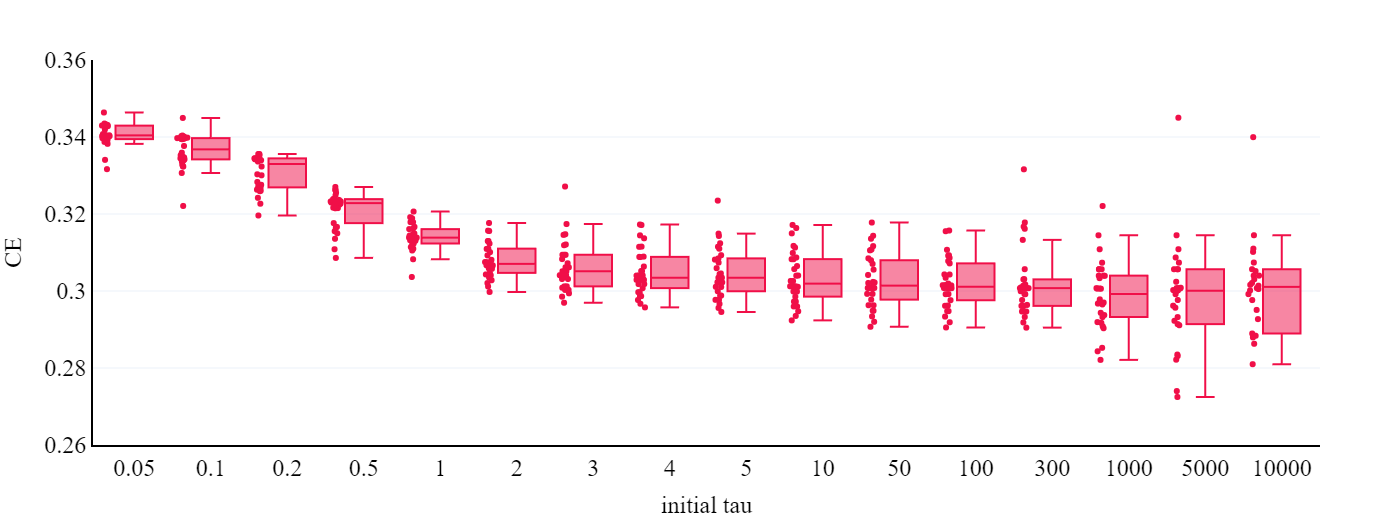}
				\caption{The medians of final CE values are different per $\tau$ values in case of the fixed $\tau$ algorithm. This figure presents the results on the Blood dataset, however, it is important that other tested classification problem show the same behaviour.}
				\label{fig:blood_allCE_avgtauind_fix}
			\end{figure}

\begin{figure}[H]
				\centering
				\includegraphics[scale=0.3]{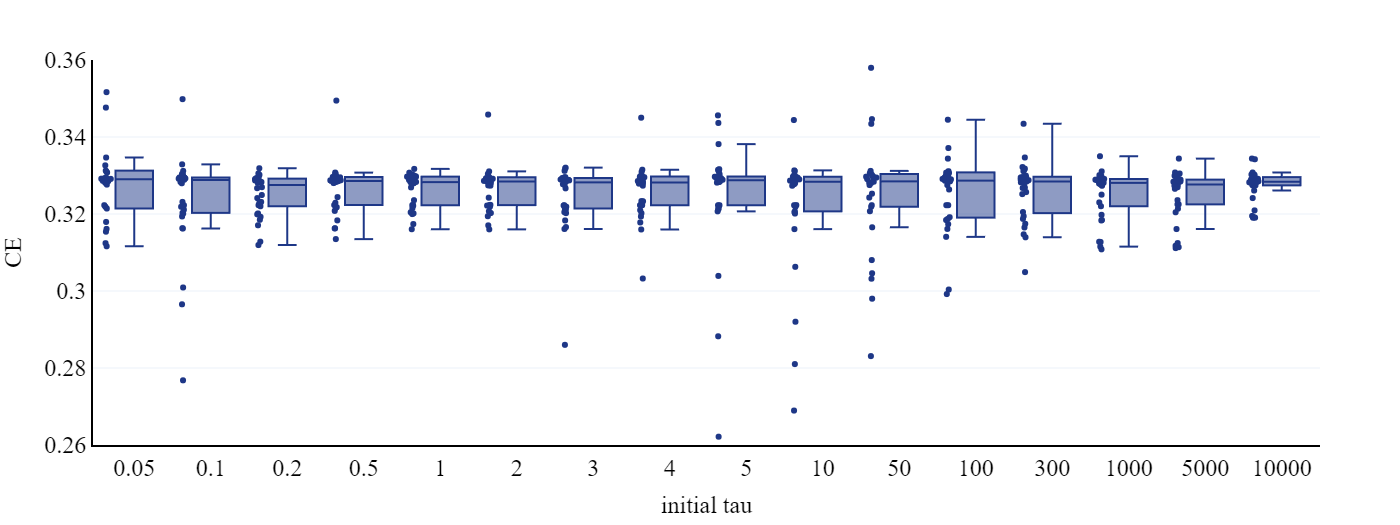}
				\caption{The medians of final CE values are similar for the tested $\tau$ values, which shows the $\tau$ independence of dynamic $\tau$ algorithm. This figure presents the results on the Blood dataset, however, it is important that other tested classification problem show the same behaviour.}
				\label{fig:blood_allCE_avgtauind_dyn}
			\end{figure}

			\textit{CE:}			
			The experienced stability is observable also in case of the CE measure. There are almost no differences among the median values for the dynamic version (Fig. \ref{fig:blood_allCE_avgtauind_dyn}) compared to the fix one (Fig. \ref{fig:blood_allCE_avgtauind_fix}). It is interesting that on the right side in the Fig. \ref{fig:blood_allCE_avgtauind_dyn} the CE values are almost the same, however, it is natural, because the high $\tau$ values realize MSE oriented learning (without considering the CE aspect).
			
			\subsubsection{The range of errors final value are independent of $\tau$} 
			The previous section \ref{same_average} dealt with the average of final error values and now their ranges are observed.
			
			\textit{$E_{ExpAbs}$:}			
			Similar independence was visible for the range of model accuracy measures for different initial $\tau$ values at the dynamic version.
			
			The range for $E_{ExpAbs}$ was similar for all $\tau$ values. This could mean that the algorithms follow a similar path, for the same initial network for different $\tau$-s.
			
			\begin{figure}[H]
				\centering
				\includegraphics[scale=0.3]{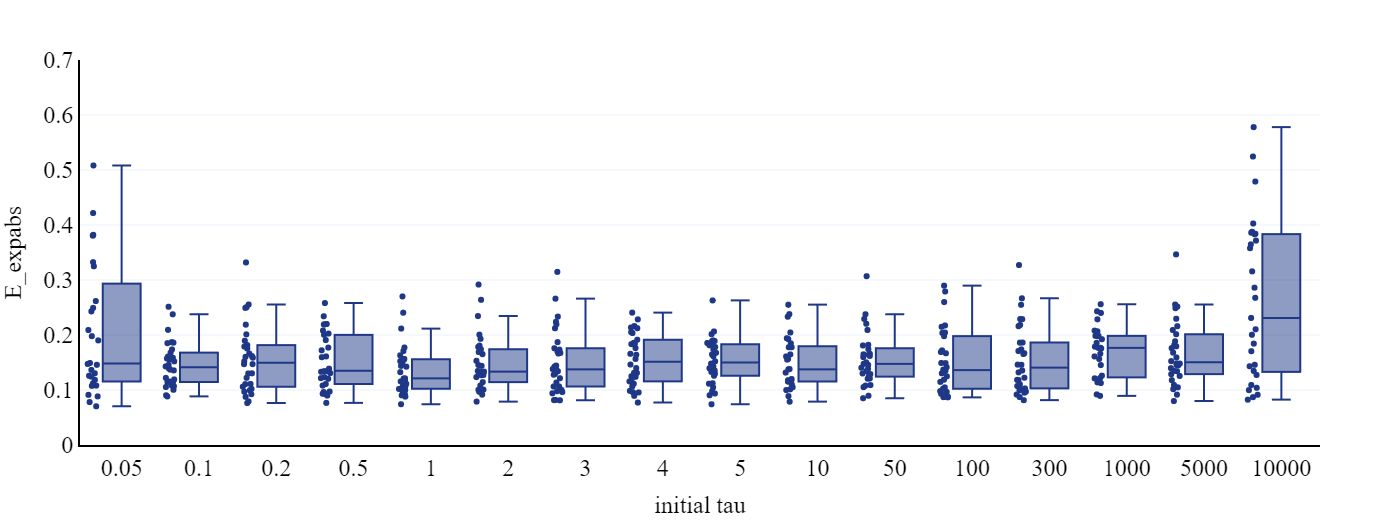}
				\caption{The range of final $E_{ExpAbs}$ values are similar for the tested $\tau$ values, which shows the $\tau$ independence of dynamic $\tau$ algorithm. This figure presents the results on the IRIS dataset, however, it is important that other tested classification problem show the same behaviour.}
				\label{fig:iris_allEexpabs_rangetauind_dyn}
			\end{figure}

			\textit{MSE:}			
			This independent behaviour could be seen for MSE, too. In case of dynamic algorithm the final MSE values are independent from the initial $\tau$-s.
			\begin{figure}[H]
				\centering
				\includegraphics[scale=0.3]{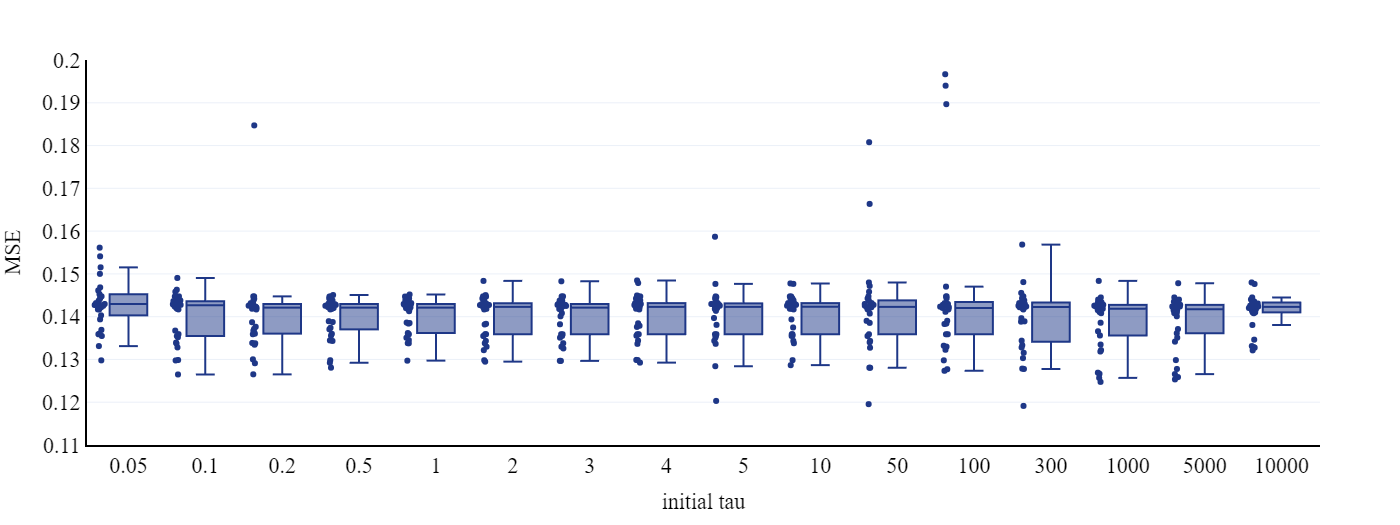}
				\caption{The range of final MSE values are similar for the tested $\tau$ values, which shows the $\tau$ independence of the dynamic $\tau$ algorithm. This figure presents the results on the Blood dataset, however, it is important that other tested classification problem show the same behaviour.}
				\label{fig:blood_allMSE_rangetauind_dyn}
			\end{figure}

			\textit{CE:}			
			For final CE the dynamic algorithm had similar $\tau$ - independent results.
			\begin{figure}[H]
				\centering
				\includegraphics[scale=0.3]{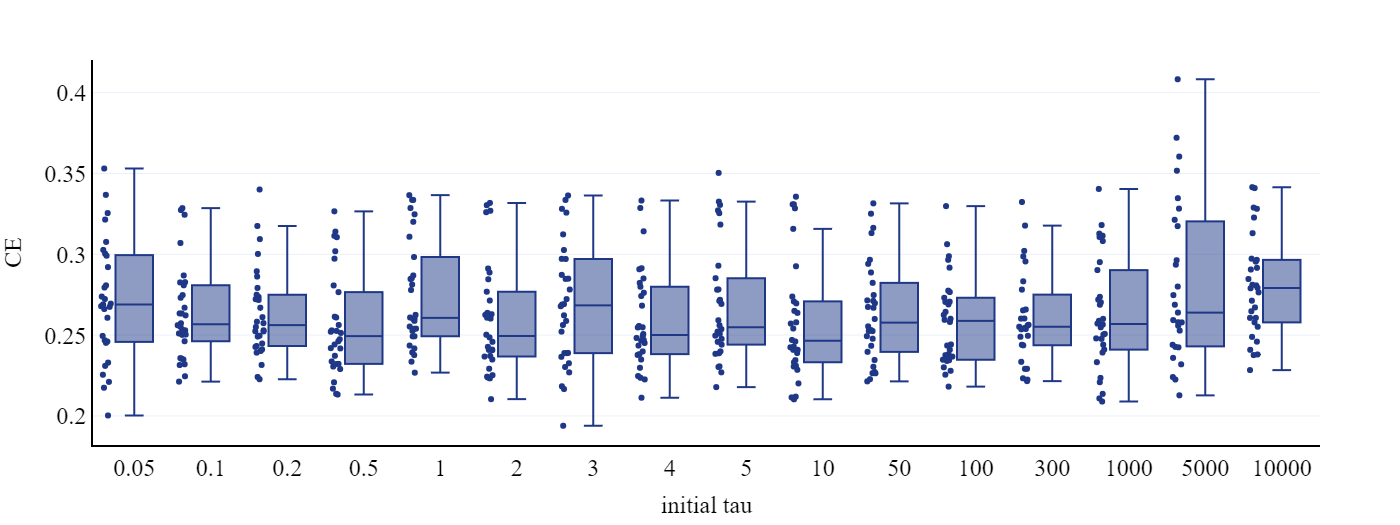}
				\caption{The range of final CE values are similar for the tested $\tau$ values, which shows the $\tau$ independence of the dynamic $\tau$ algorithm. This figure presents the results on the Thyroid dataset, however, it is important that other tested classification problem show the same behaviour.}
				\label{fig:thyroid_allCE_rangetauind_dyn}
			\end{figure}
			
			\subsubsection{The average training step numbers for different initial $\tau$ values are the same}
			
\begin{figure}[H]
				\centering
				\includegraphics[scale=0.3]{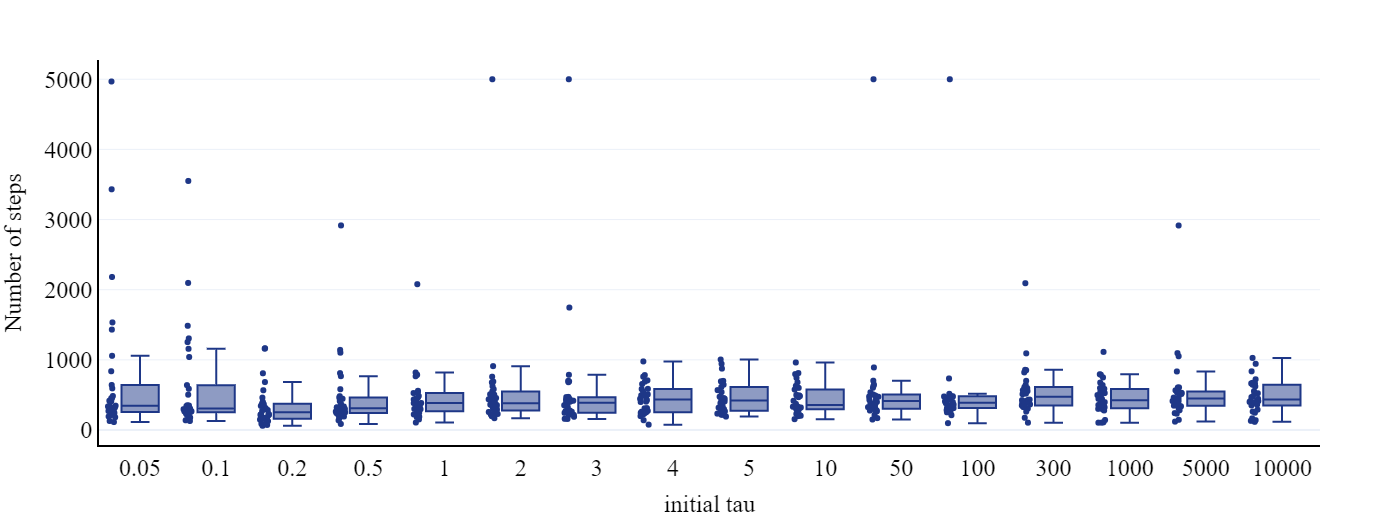}
				\caption{The box plot of the 30 required iteration step numbers are very similar for the tested initial $\tau$ values. This figure presents the results on the Ionosphere dataset, however, it is important that other tested classification problem show the same behaviour.}
				\label{fig:ionosphere_allStep_rangetauind_dyn}
			\end{figure}

			\textit{Number of training steps:}			
			The $\tau$ independence was visible not only in the case of model accuracy measures, but in the required iteration step numbers as well (Fig. \ref{fig:ionosphere_allStep_rangetauind_dyn}). It means also that the runtime of the proposed, novel, dynamic algorithm is not dependent on the initial values of the dynamically varying $\tau$ parameter.
			
			\subsubsection{Experiments on regression (non-classification) datasets} \label{nonclass}
			
			Most of benefits are not observable in case of non-classification problems, similarly to \cite{silva2008data}.
			
			\begin{figure}[H]
				\centering
				\includegraphics[scale=0.3]{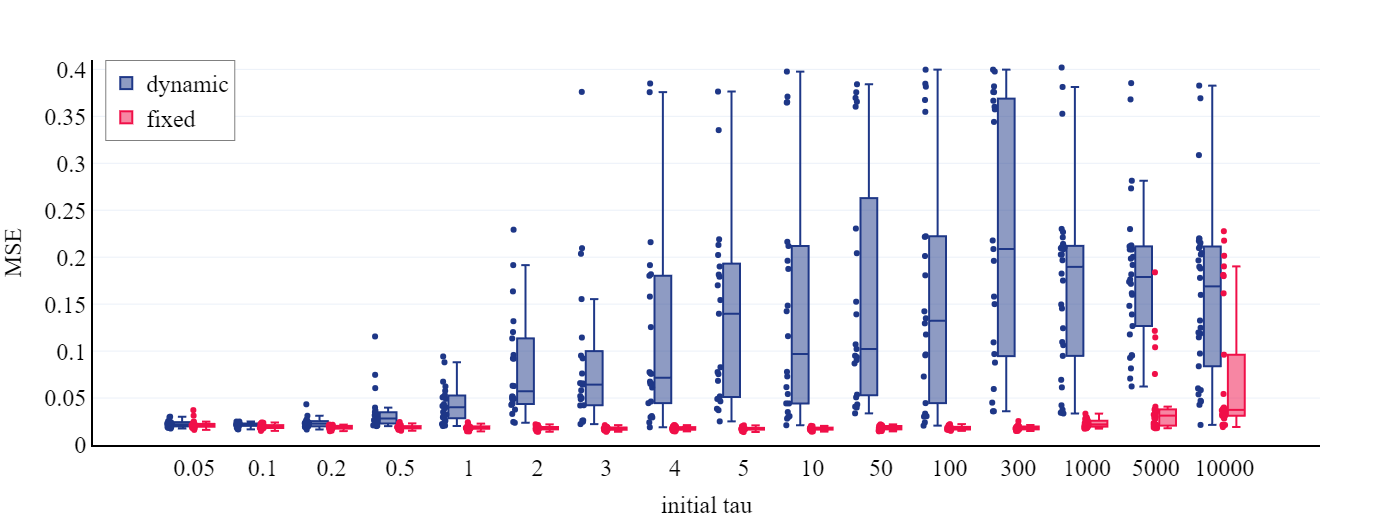}
				\caption{MSE values on a regression dataset Diabetes, it does not show the benefits mentioned in the previous sections.}
				\label{fig:diab_regression_MSE}
			\end{figure}
		
			As it is shown in figure \ref{fig:diab_regression_MSE}, the final MSE values of the dynamic $\tau$ algorithm did not reach better results than the fixed one in case of a regression datasets (e.g. Diabetes), however, for higher $\tau$ values its ranges seems to be similar, so it may have some advantages as well, but, this field could be a challenge of further research analysis.
			
			In this research field \cite{heravi2018comparison, heravi2018does, heravi2018new, heravi2019new} presented excellent research results analysing noise and other features inherited in the training datasets to predict various properties of the resulted training models.

	    \subsection{Superior model accuracy over the best performing state-of-the-art combined error measure based training by  \cite{amaral2013using}} \label{adult_res}
	    
	    %
	    
\begin{figure}[H]
			\centering
			\includegraphics[scale=0.6]{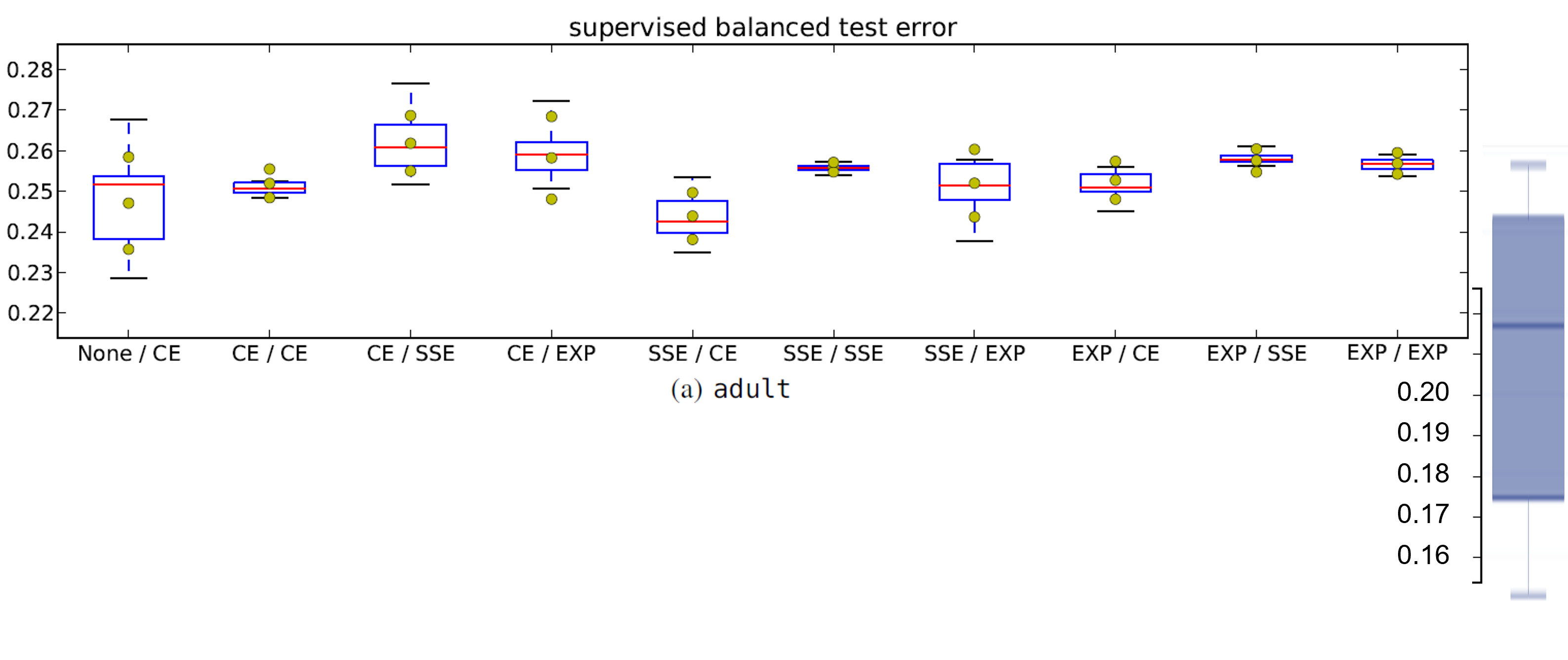}
			\caption{Recognition test errors (1 - Recognition Rate) of the new algorithm (blue boxplot on the right) compared to result of \cite{amaral2013using} (white boxplots on the left) showing that the proposed, novel, dynamic algorithm can reach significantly better performance in case of 2 hidden layers with 2-2 hidden nodes.}
			\label{fig:adult_recrate22}
		\end{figure}

\begin{figure}[H]
			\centering
			\includegraphics[scale=0.37]{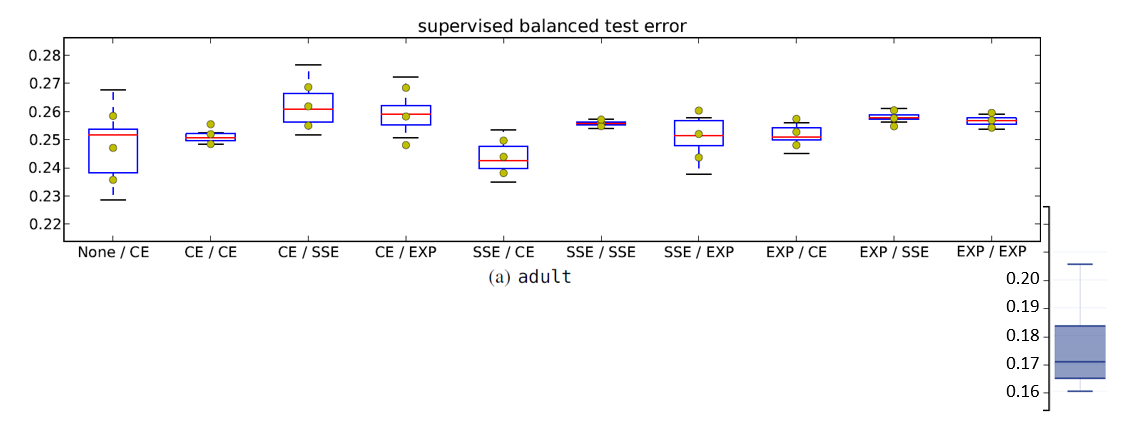}
			\caption{Recognition test errors (1 - Recognition Rate) of the new algorithm (blue boxplot on the right) compared to result of \cite{amaral2013using} (white boxplots on the left) showing that the proposed, novel, dynamic algorithm can reach significantly better performance in case of 1 hidden layer with 8 hidden nodes.}
			\label{fig:adult_recrate8}
		\end{figure}

	    The section \ref{adult_intro} described that the new training algorithm was compared to state-of-the-art research result applying combined error measures. 
	    \cite{amaral2013using} analysed the behaviour of auto-encoders trained by different cost functions. They used in the pre-training and fine-tuning phase two different cost functions like cross-entropy (CE), sum of squared error (SSE) and exponential error (EXP). The classification experiments showed that the SSE is the best for pre-training. It was observed that CE function in fine-tuning treats well the unbalanced data. Consequently, the best pair of pre-training and fine-tuning is SSE and CE (SSE/CE in Fig. \ref{fig:adult_recrate22} and \ref{fig:adult_recrate8}) 
	    
	    This comparison showed that, \textit{the proposed, new, full dynamic method served with much more accurate models than the presented algorithm by \cite{amaral2013using} realizing a sequentially applied, two steps learning approach}.
	
	\section{Conclusions}
		It is well-known that artificial neural network based modelling is a promising, continuously evolving field of artificial intelligence. The variety of applied error measures in the training/validation process motivated the research and the related comprehensive literature review identified that that there is no "best of" measure, although there is a set of measures with changing superiorities in different learning situations. Consequently, the combinations and/or competition of the available measures may serve with advances to improve the learning. In this relation an outstanding, remarkable measure called $E_{Exp}$ published by Silva and his research partners (\cite{silva2008data, silva2014classification})  represents a research direction to combine more measures successfully with fixed importance weighting (determined by the parameter $\tau$) during learning, it is the actual state-of-the-art according to the aspect of training error combinations. Their recent results serves also with a two, consecutive step solution by \citeauthor{amaral2013using} of training auto-encoders by different cost functions (SSE, CE and EXP), they used in the pre-training and fine-tuning phase two different cost functions of them. The classification experiments showed that the SSE is the best for pre-training and CE function in fine-tuning, consequently, the best pair of pre-training/fine-tuning is SSE/CE.
		
		\textbf{The main idea of the paper is} to go far beyond and to \textbf{integrate the relative importance introduced by Silva and his colleagues into the neural network training algorithm(s) realized through a novel error measure called $E_{ExpAbs}$, moreover, the proposed solution is in the same time a dynamic training algorithm, so, no consecutive, separate training stages are needed}. This approach is included into the (accelerated) Levenberg-Marquardt training algorithm, so, \textbf{a novel version of Levenberg-Marquardt training was also introduced, resulting in a self-adaptive, dynamic learning algorithm}. 
		This dynamism does not has positive effects on the resulted model accuracy only, but also on the training process itself. \textbf{The described comprehensive algorithm tests proved that it integrates \textit{dynamically} the two big worlds of statistics and information theory that is the key novelty of the paper}.
		
		The introduced, novel, dynamic algorithm was comprehensively tested on 13 well-known machine learning dataset and it was always compared to the original, fixed importance weighting version (so, to the original, not dynamic version), moreover, the algorithm capability was also directly compared to the most recent, two-stage learning solution as well. \textbf{As result, the proposed, novel, dynamic training algorithm shows a number of superior features over the actually available, integrated training algorithms}:
		\begin{itemize}
			\item All of the superior features of the proposed algorithm are valid for neural network model Mean-Squared Error ($MSE$), Cross-Entropy ($CE$), for the novel, introduced model error measure $E_{ExpAbs}$ and for the Recognition Rate ($RR$) as well, consequently, \textbf{it shows advances from various viewpoints of statistics and information theory independently}.
			\item For smaller and also for larger initial $\tau$ values, the dynamic algorithm reached superior results, than the fixed valued algorithm. It is a very important feature, because, when using the original, fixed $\tau$ version this parameter has to be determined before starting the training process, moreover, as it was well presented in the related publications, many test had to be performed to find the most suitable value for it. \textbf{With the novel, proposed, dynamic algorithm none of these pre-tuning steps are needed}.
			\item The average resulted model accuracy measures for different initial $\tau$ values are the same, moreover it is valid for the most widely applied model error measures of $MSE$, $CE$ and $RR$. \textbf{The related tests mirrored that without the preliminary knowledge about the suitable $\tau$ it can tune it well during the training and can fine the appropriate value for it. This feature is especially important because it realizes a learning technique that tunes one of its key training parameters according to the specialities of the given training dataset. Additionally, the final training measure $E_{ExpAbs}$, is in a much smaller range than with the fixed version after the training, so, the proposed algorithm has higher robustness as well}. 
			\item \textbf{The introduced, speed increase technique} (combined Levenberg-Marquardt, 'Momentum' and 'SuperSAB' steps) \textbf{resulted in a convergent solution, moreover, the such embedded, dynamic algorithm is multiple times quicker than the original, fixed $\tau$ version}. Please, note that the fixed $\tau$ version does not allow any acceleration because this parameter cannot be changed during training. This acceleration resulted also that the number of training steps are the same, independent from the starting (initial) $\tau$ \textbf{that is an important robustness behaviour concerning the learning process itself}.
			\item The state-of-the-art, combined error measure technique introduced by \cite{amaral2013using} already divides the training into two consecutive stages, the introduced, dynamic solution realizes it in each learning iteration, so, \textbf{it resulted significant} (event the evaluation range had to be extended) \textbf{superiority in the analysed recognition range of the given dataset}.
		\end{itemize}
		
		The proposed, novel dynamic training algorithm serves with various superiority behaviour but it is important to note that \textit{these results are valid only for the classification, but not for the regression assignments}, according to the various assessments. Probably this is because the introduced $E_{ExpAbs}$ measure combines CE with MSE and CE has advances typically in classification tasks, but this is only an assumption that needs further research.
	\section{Outlook}
        During the research progress the authors continuously collected novel research ideas as open challenges:
        \begin{itemize}
				\item The introduced $E_{ExpAbs}$ measure integrates $MSE$ and $CE$ and has many advantageous features. However, in the literature there are many other possible measures (e.g. Rényi Entropy) as it was introduced at the beginning of the paper. To integrate more measures into a novel one could be a promising direction for further research. Ideal would be having a "super-measure" that integrates (almost) all of the beneficial measures into one error "calculus". A competitive training by several measures could be an alternative, similar to \cite{viharos_adaptive_2021} in case of feature selection. As it was introduced and compared above, \cite{amaral2013using} combined sequentially more error measures and found an optimal sequence of them. A potential continuation of this competitive training research could be to vary the fixed $\tau$ version and the dynamic $\tau$ version.
				\item The proposed algorithm is a novel version of the Levenberg-Marquardt (LM) training algorithm, however, the original solution has two different kinds. The current one applies a smaller Jacobian matrix introduced by \cite{heravi2016new}, its size is Number of network weights x Pattern number. However, there exists another version having a larger Jacobian matrix introduced by \cite{HaoYu.2011}, in this case the matrix's size is Number of network weights x Pattern number x Network output number. All the introduced extension steps and experiments can be realized also with the "large matrix" based solution in the future, the first solutions in this direction were already developed by the authors of this paper.
				\item There are more recent and more popular training algorithms than LM, e.g. ADAM is one of the most frequently applied one (\cite{kingma_adam_2017}). The application of the introduced $E_{ExpAbs}$ measure in the ADAM (or other) training algorithm is also one of its possible extensions.. Actually, the authors already developed the first working version of this novel algorithm included in ADAM learning - it will be the topic of a next publication.
				\item The proposed, adaptive $\tau$ is a key parameter for the training process, however, there is a potential to introduce also dynamic but different $\tau$ parameters for each of the individual neural network outputs. The theoretical basis of this direction was already prepared by the authors.
				\item The mixed training error measure $E_{ExpAbs}$ is promising for realizing a better feature selection algorithm inside the so called AHFS algorithm by \cite{viharos_adaptive_2021}.
		\end{itemize}
        
	\section{Acknowledgement}
	    This research has been supported by the TKP2021-NKTA-01  NRDIO grant on "Research on cooperative production and logistics systems to support a competitive and sustainable economy" (COPROLOGS) and by the European Union project RRF-2.3.1-21-2022-00004 within the framework of the Artificial Intelligence National Laboratory.
	    
	    On behalf of Project "Comprehensive Testing of Machine Learning Algorithms" we are grateful for the usage of ELKH Cloud (see \cite{H_der_2022}; https://science-cloud.hu/) which helped us achieve the results published in this paper.
	    
	\appendix
	\section{Elements of $E_{ExpAbs}$ derivatives - detailed computation} \label{E_expabs_derivatives_appendix}
	
	\begin{equation}
		\begin{aligned}
			\frac{\partial err_p^{ExpAbs}}{\partial \tau}
			&= sign(\tau) \cdot \exp{\left( \frac1{|\tau|} \sum_{n_L = 0}^{N_L} e_{n_L}^2\right)}\\ 
			&+ |\tau| \cdot \exp{\left( \frac1{|\tau|} \sum_{n_L = 0}^{N_L} e_{n_L}^2 \right)} \cdot \left( -1 \right) \cdot \frac1{{|\tau|}^2} \ sign(\tau) \left( \sum_{n_L = 0}^{N_L} e_{n_L}^2 \right) \\
			&= sign(\tau) \cdot \exp{\left( \frac1{|\tau|} \sum_{n_L = 0}^{N_L} e_{n_L,p}^2 \right)} \cdot \left( 1 - \frac1{|\tau|} \sum_{n_L = 0}^{N_L} e_{n_L,p}^2 \right) \\
			&= sign(\tau) \exp{\left( \frac1{|\tau|} \sum_{n_L = 0}^{N_L} {\left( t_{n_L} - o_{n_L}^L \right)}^2 \right)} \cdot \left( 1 - \frac1{|\tau|} \sum_{n_L = 0}^{N_L} {\left( t_{n_L} - o_{n_L}^L \right)}^2 \right)
		\end{aligned}
	\end{equation}
	
	\begin{equation}
		\begin{aligned}
			\frac{\partial{err_p^{Exp}}}{\partial{o_{n_L}^L}}
			&= |\tau| \exp{ \left( \frac1{|\tau|} \sum_{{n_L}=0}^{N_L} \left( t_{n_L} - o_{n_L}^L \right) ^2 \right) } \cdot \frac1{|\tau|} \cdot 2 \cdot \left( t_{n_L} - o_{n_L}^L \right) \cdot \left( -1 \right) \\
			&= - 2 \cdot \left[ \exp{ \left( \frac1{|\tau|} \sum_{{n_L}=0}^{N_L} \left( t_{n_L} - o_{n_L}^L \right) ^2 \right) } \right] \cdot \left( t_{n_L} - o_{n_L}^L \right) \\
			&= - 2 \cdot \left[ \exp{ \left( \frac1{|\tau|} \sum_{{n_L}=0}^{N_L} e_{n_L,p} ^2 \right) } \right] \cdot e_{n_L,p}
		\end{aligned}
	\end{equation}
	
	\begin{equation}
		\begin{aligned}
			\frac{ \partial{o_{n_L}^L} }{ \partial{o_{n_{L-1}}^{L-1}}}
			&= \frac{ \partial }{ \partial{o_{n_{L-1}}^{L-1}}} \left( \sigma \left( \sum_{n_{L-1}=0}^{N_{L-1}} w_{n_{L-1},n_L}^L \cdot o_{n_{L-1}}^{L-1} \right) \right) \\
			&= \sigma ' \left( \sum_{n_{L-1}=0}^{N_{L-1}} w_{n_{L-1},n_L}^L \cdot o_{n_{L-1}}^{L-1} \right) \cdot \frac{ \partial }{ \partial{o_{n_{L-1}}^{L-1}}} \left( \sum_{n_{L-1}=0}^{N_{L-1}} w_{n_{L-1},n_L}^L \cdot o_{n_{L-1}}^{L-1} \right) \\
			&= \sigma ' \left( \sum_{n_{L-1}=0}^{N_{L-1}} w_{n_{L-1},n_L}^L \cdot o_{n_{L-1}}^{L-1} \right) \cdot w_{n_{L-1},n_L}^L \\
			&= \sigma \left( \sum_{n_{L-1}=0}^{N_{L-1}} w_{n_{L-1},n_L}^L \cdot o_{n_{L-1}}^{L-1} \right) \cdot \left( 1 - \sigma \left( \sum_{n_{L-1}=0}^{N_{L-1}} w_{n_{L-1},n_L}^L \cdot o_{n_{L-1}}^{L-1} \right) \right) \cdot w_{n_{L-1},n_L}^L \\
			&= \sigma{ \left( o_{n_L}^L \right) } \cdot \left( 1 - \sigma{ \left( o_{n_L}^L \right) } \right) \cdot w_{n_{L-1},n_L}^L
		\end{aligned}
	\end{equation}

	\begin{equation}
		\begin{aligned}
			\frac{ \partial{o_{n_l}^l} }{ \partial{o_{n_{l-1}}^{l-1}}}
			&= \frac{ \partial }{ \partial{o_{n_{l-1}}^{l-1}}} \left( \sigma \left( \sum_{n_{l-1}=0}^{N_{l-1}} w_{n_{l-1},n_l}^l \cdot o_{n_{l-1}}^{l-1} \right) \right) \\
			&= \sigma ' \left( \sum_{n_{l-1}=0}^{N_{l-1}} w_{n_{l-1},n_l}^l \cdot o_{n_{l-1}}^{l-1} \right) \cdot \frac{ \partial }{ \partial{o_{n_{l-1}}^{l-1}}} \left( \sum_{n_{l-1}=0}^{N_{l-1}} w_{n_{l-1},n_l}^l \cdot o_{n_{l-1}}^{l-1} \right) \\
			&= \sigma ' \left( \sum_{n_{l-1}=0}^{N_{l-1}} w_{n_{l-1},n_l}^l \cdot o_{n_{l-1}}^{l-1} \right) \cdot w_{n_{l-1},n_l}^l \\
			&= \sigma \left( \sum_{n_{l-1}=0}^{N_{l-1}} w_{n_{l-1},n_l}^l \cdot o_{n_{l-1}}^{l-1} \right) \cdot \left( 1 - \sigma \left( \sum_{n_{l-1}=0}^{N_{l-1}} w_{n_{l-1},n_l}^l \cdot o_{n_{l-1}}^{l-1} \right) \right) \cdot w_{n_{l-1},n_l}^l \\
			&= \sigma{ \left( o_{n_l}^l \right) } \cdot \left( 1 - \sigma{ \left( o_{n_l}^l \right) } \right) \cdot w_{n_{l-1},n_l}^l
		\end{aligned}
	\end{equation}
	
	\begin{equation}
		\begin{aligned}
			\frac{ \partial{o_{n_l}^l} }{ \partial{w_{n_{l-1},n_l}^{l}}}
			&= \frac{ \partial }{ \partial{w_{n_{l-1},n_l}^{l}}} \left( \sigma \left( \sum_{n_{l-1}=0}^{N_{l-1}} w_{n_{l-1},n_l}^l \cdot o_{n_{l-1}}^{l-1} \right) \right) \\
			&= \sigma ' \left( \sum_{n_{l-1}=0}^{N_{l-1}} w_{n_{l-1},n_l}^l \cdot o_{n_{l-1}}^{l-1} \right) \cdot \frac{ \partial }{ \partial{w_{n_{l-1},n_l}^{l}}} \left( \sum_{n_{l-1}=0}^{N_{l-1}} w_{n_{l-1},n_l}^l \cdot o_{n_{l-1}}^{l-1} \right) \\
			&= \sigma ' \left( \sum_{n_{l-1}=0}^{N_{l-1}} w_{n_{l-1},n_l}^l \cdot o_{n_{l-1}}^{l-1} \right) \cdot o_{n_{l-1}}^{l-1} \\
			&= \sigma \left( \sum_{n_{l-1}=0}^{N_{l-1}} w_{n_{l-1},n_l}^l \cdot o_{n_{l-1}}^{l-1} \right) \left(1 - \sigma \left( \sum_{n_{l-1}=0}^{N_{l-1}} w_{n_{l-1},n_l}^l \cdot o_{n_{l-1}}^{l-1} \right) \right) \cdot o_{n_{l-1}}^{l-1} \\
			&=  \sigma{ \left( o_{n_l}^l \right) } \cdot \left( 1 - \sigma{ \left( o_{n_l}^l \right) } \right) \cdot o_{n_{l-1}}^{l-1}
		\end{aligned}
	\end{equation}

	\bibliographystyle{model5-names} 
	\biboptions{authoryear}
	\bibliography{./LM_Eexpabs_references_20200721}

	\end{document}